\documentclass{article}

\usepackage{PRIMEarxiv}
\usepackage{amsmath}
\usepackage[utf8]{inputenc} % allow utf-8 input
\usepackage[T1]{fontenc}    % use 8-bit T1 fonts
\usepackage{hyperref}       % hyperlinks
\usepackage{url}            % simple URL typesetting
\usepackage{booktabs}       % professional-quality tables
\usepackage{amsfonts}       % blackboard math symbols
\usepackage{nicefrac}       % compact symbols for 1/2, etc.
\usepackage{microtype}      % microtypography
\usepackage{lipsum}
\usepackage{algorithm}
\usepackage{algpseudocode}
\usepackage{fancyhdr}       % header
\usepackage{graphicx}       % graphics
\graphicspath{{Figs/}}     % organize your images and other figures under Figs/ folder

\title{Curvature Augmented Manifold Embedding and Learning
\thanks{\textit{\underline{Citation}}: 
\textbf{TBD}} 
}

\author{
  Yongming Liu\\
  Mechanical and Aerospace Engineering\\
  Arizona State University \\
  Tempe, AZ USA\\
  \texttt{\ yongming.liu@asu.edu} \\
}

\begin{document}
\maketitle

\begin{abstract}
%% Text of abstract
A new dimensional reduction (DR) and data visualization method, Curvature-Augmented Manifold Embedding and Learning (CAMEL), is proposed. The key novel contribution is to formulate the DR problem as a mechanistic/physics model, where the force field among nodes (data points) is used to find an n-dimensional manifold representation of the data sets. Compared with many existing attractive-repulsive force-based methods, one unique contribution of the proposed method is to include a non-pairwise force. This formulation is inspired by the multi-body potential in physics and curvature in topology. The paper first gives a brief review of existing DR methods. It focuses on a unified view of many different formulations in terms of the analogy of gradient computing with an attractive-repulsive force field. Following this, a new force field model is introduced and discussed, along with the background of multi-body potential in lattice-particle physics and Riemann curvature in topology. A curvature-augmented force is included in CAMEL computing. Following this, CAMEL formulation for unsupervised learning, supervised learning, semi-supervised learning/metric learning, and inverse generation are provided. Next, CAMEL is applied to many benchmark datasets by comparing existing models, such as tSNE, UMAP, TRIMAP, and PacMap. Both visual comparison and metrics-based evaluation are performed. 14 open literature and self-proposed metrics are employed for a comprehensive comparison. Finally, conclusions and future work are suggested based on the current investigation. Related code and demonstration are available on https://github.com/ymlasu/CAMEL for interested readers to reproduce the results and other applications.

\end{abstract}

% keywords can be removed

\section{Introduction}
Dimension reduction (DR) is a long-lasting and focused area in engineering, science, and machine learning communities. It may have different names and preferences depending on the individual field. For example, in engineering, it can referred to as reduced-order modeling, and it is closely related to data visualization in machine learning. The core concept is to solve the curse of dimensionality by projecting the data features to a low dimensional space (2D or 3D for data visualization problems, but not necessary for general DR problems). Once the low-dimensional data structure is obtained, many analyses, such as classification and regression, can be done conveniently compared to their counterparts in the high-dimensional spaces.

The DR method can be traced back to the most widely used principal component analysis (PCA) \cite{jolliffe2016principal}, a linear DR method based on the eigenvalue problems of all data points. PCA has alternative names in engineering and science, such as proper orthogonal decomposition \cite{kerschen2005method} in structural dynamics and Kahunen-Leove expansion in engineering statistics\cite{betz2014numerical}. The nonlinear DR method has been proposed to improve the apparent limitation of the linear DR method, such as locally linear embedding (LLE)\cite{roweis2000nonlinear}, ISOMAP\cite{balasubramanian2002isomap}, and Laplacian Eignemap\cite{belkin2003laplacian}, among many others. A detailed review of these earlier developments can be found in \cite{van2009dimensionality}. One important and interesting category for the nonlinear DR method is the force-directed graph method \cite{kobourov2012spring}, where the forces among nodes in a graph are used to find the visualization in low dimension. The reason to have a special mention here is that it uses the concept of "force" to guide the final representation of data, which is later shown that the concept of "force" can be used to unify many DR methods, although they have been initially proposed by different conceptualization.

Stochastic Neighboring Embedding (SNE)\cite{roweis2000nonlinear} and its t-distributed version (t-SNE)\cite{van2008visualizing} is one of the most critical breakthroughs in the nonlinear DR. The key concept is to define the probability with neighboring and distant points, where the low-dimensional presentation (2D or 3D) tries to reproduce this probabilistic similarity measure. The original version of t-SNE does not scale well with the number of samples, and the improved version using the tree-based algorithm is proposed \cite{van2014accelerating}. Later, an alternative approach (LargeVis) is proposed to improve scalability significantly with larger datasets\cite{tang2016visualizing}. A separate definition of the loss function concerning neighboring and distant points, together with negative sampling, is used to reduce the computational complexity. A Riemannian manifold-based nonlinear DR method, named Uniform Manifold Approximation and Projection (UMAP)\cite{mcinnes2018umap}, is proposed. UMAP is inspired by the fuzzy topological representation of the high-dimensional data and tries to optimize the low-dimensional representation with a defined loss function. The loss function has a similar separation to that in LargeVis (for neighboring and distant points) and implements the negative sampling to speed up computation. It is one of the most cited DR methods in recent years, partially due to the well-documented software package and tutorials. The current author also greatly benefits from the UMAP implementation in developing this work. A new method using the concept of triplets (e.g., three points rather than a pair of points) to do the DR is proposed and is named TRIMAP \cite{amid2019trimap}. The triplet is also divided into a neighboring point and a distant point, which shares a similar concept with the above-mentioned separation of data points depending on the distance. A following development, named PaCMAP \cite{wang2021understanding}, is proposed using both neighboring and distant points, together with a so-called mid-near point, where the point is typically between the nearest neighbors and distant points. This can be considered as an additional tuning for the DR behavior. It was later observed that the added mid-near point has minimal impact on the embedding \cite{jimelvilleNotesonPACMAP}. The software framework by TRIMAP/PaCMAP is very compact and easy to re-develop, which facilitates many continued work and use cases. The current model's implementation is also based on the same structure.

Despite the numerous models proposed in the DR, particularly the t-SNE type DR method, the existing method has many common characteristics/disadvantages. First, almost all methods are highly sensitive to the model parameter tuning. This tuning depends on the number of data points, the number of neighboring points in the k-nearest network (kNN) diagram, and the subset sampling ratio in some algorithms (e.g., negative sampling ratio in LargeVis, UMAP and the far-to-near ratio in PaCMAP). Thus, the embedding and projection is a trial-and-error method for some applications. Next, many existing methods use weighting tricks during the optimization process to stabilize the DR. For example, t-SNE uses early exaggeration to increase the weights of neighboring points. UMAP uses iteration-dependent weighting, and PaCMAP defines three-stage weighting factors. These tricks are understandable from an optimization algorithmic point of view, but the determination of the tricks is arbitrary, and there is no mechanism to interpret and guide these iteration-dependent tricks. Finally, most existing algorithms heavily focused on unsupervised high-to-low dimensional projection (usually 2D and 3D). Exploration for other tasks (such as supervised and semi-supervised learning) is lacking. Specifically, most methods are designed for high-to-low DR, and the inverse projection from low-to-high dimensions is rarely touched. UMAP is the only one that provides a complete solution, which is one reason that it has had extensive usage in recent years, as applications may need very different functionalities of the DR method. One major motivation of the proposed method is to develop a compact, extensible, and unified (high-low and low-high) DR framework for very different applications.

Another critical and interesting observation of all these methods is that they are all very similar and can be unified! Many recent studies tried to compare different DR methods and observed that different methods could yield virtually identical results if some parameters can be adjusted \cite{bohm2022attraction}. The gradient patterns of several DR methods are analyzed and compared \cite{wang2021understanding}, and it is shown empirically that they share very similar patterns for neighboring and distant points. A recent study compares t-SNE and UMAP using contrastive learning and shows that they are very similar if negative sampling is considered \cite{damrich2022t}. Some analytical studies have been developed to compare the gradient expression of different methods, e.g., t-SNE, UMAP, LargeVis, and PaCMAP in \cite{jimelvilleSomeTheory,jimelvilleNotesonPACMAP}. It was observed that the form of gradients is very similar if they can be written into positive and negative components. These positive-negative gradients are expressed analogically as the attractive-repulsive force \cite{bohm2022attraction}. t-SNE, UMAP, Laplacian Eigenmap, and another force-directed graph method (ForceAtlas2\cite{jacomy2014forceatlas2}) are unified in a force-spectrum plot, where each of them has a different attraction/repulsion ratio. Laplacian Eigenmap, a well-known method for its capability to highlight the local structure, has the largest attraction/repulsion ratio. This high attraction/repulsion ratio tends to have densely packed clusters. t-SNE, on the other hand, has the smallest attraction/repulsion ratio, which leads to more uniformly formed clusters. UMAP and ForceAtlas2 are somewhere in between these two extreme cases. It is also shown that this behavior is highly correlated with the parameter settings. The weight factor in the repulsive force term is found to be related to the number of data points, neighbor numbers, and negative sampling rate. The above review suggests that many different DR methods regarding the force field analogy to the classical mechanics/physics problem can be unified. Thus, many available models and knowledge in mechanics/physics can be used to enhance the DR method. In addition, since these models are based on certain mechanisms, the model parameters (specifically the weight factors of different forces) are interpretable. This may pave the road to unifying parameter calibration with mechanism guidance. This is the major motivation of the proposed study.

The remaining part of this paper is organized as follows. First, related work is briefly reviewed, and the focus is on highlighting the similarities and differences to unify the expression of the t-SNE-type DR method. Next, a proposal to use the multi-body force to augment the existing force (gradients) models is presented. This is inspired by the multi-body potential in physics and is also closely related to the Riemann curvature in topology. Detailed derivation for the force (gradients) of the proposed CAMEL is given. Algorithmic details are also provided for different learning tasks, such as unsupervised, supervised, metric, and inverse learning. Following this, a demonstration using various benchmark examples is shown, together with a comparison of several existing DR methods. Discussions on several impacts of the proposed model, such as curvature-induced force, the robustness of the parameter model, scalability, supervision, and interpretability, are presented. Finally, several conclusions and future work are presented based on the current investigation.

\section{Related work}
\label{sec:headings}
A brief review of related work is presented here. As we focus on the DR method that is optimized by the gradient-based method, other types of DR methods are not reviewed here, for example, the optimization-free graph method \cite{sarfraz2022hierarchical}. In addition, extensive studies have been done to compare the pros and cons, accuracy and efficiency, and scalability of different methods. Thus, the author has no intent to repeat these reviews and interested readers are suggested to read several review articles \cite{van2009dimensionality,anowar2021conceptual,ray2021various, jolliffe2016principal, zebari2020comprehensive}. The focus of this section is to highlight the similarities and differences of various DR methods when they can be expressed in a unified force field expression, specifically,
\begin{equation}
Gradients={\frac{\partial G}{\partial y_{i}}}= Force Field = \sum_{i=1}^{N}(F_{attractive}(y_{p=1}^{N})-F_{repulsive}(y_{q=1}^{N}))
\end{equation}

where $G$ is the loss function and $Y_{i}$ is the coordinates of the $i^{th}$ the projected space. $N$ is the total number of data points. $F_{attractive}(Y_{p=1}^{N})$ and $F_{repulsive}(Y_{q=1}^{N})$ are the attractive force field and repulsive force field, respectively. $p$ and $q$ are used to indicate that all points could potentially contribute to the force on the $i^{th}$ point. It should be noted that each model has a different specific force field form. In addition, it may not be necessary (and computationally feasible for a large number of $N$) to consider all data points to determine the force field. One common approach is to use a k-nearest neighbor algorithm to separate the data into neighboring and distant points. The attractive force is dominant in neighboring points, and the repulsive force dominates in the distant point. Thus, an approximation using kNN can be written as:

\begin{equation}
Gradients={\frac{\partial G}{\partial y_{i}}}= Force Field = \sum_{p=1}^{k}(F_{attractive}(y_{p=1}^{k}))-\sum_{p=k+1}^{N}(F_{repulsive}(y_{p=k+1}^{N}))
\end{equation}
where $k$ is the number of nearest neighbor points. This approximation is almost used in every reviewed DR method to make the computation tractable.

\subsection{Method: t-SNE}
The key concept of t-SNE is to define a similarity measure in high dimension ($X$) and low dimension ($y$). The low dimension embedding is gradually adjusted so the difference between the similarity measure is minimized by minimizing the Kullback–Leibler (KL)
divergence \cite{van2008visualizing}. The similarity measure in high dimension is defined as
\begin{equation}
{p_{ij}}={\frac{v_{ij}}{N}},  {v_{ij}}={\frac{p_{i|j}+p_{j|i}}{2}},  {p_{j|i}}={\frac{v_{j|i}}{\sum_{l\neq i}{v_{l|i}}}}, v_{j|i}=exp(-\frac{||X_{i}-X_{j}||^{2}}{2\sigma _{i}^{2}})
\end{equation}

where $v_{j|i}$ is Gaussian distance measure of the point $i$ and $j$. By normalizing this distance measure with all neighbors and taking the symmetrical averaging of  $p_{i|j}$ and $p_{j|i}$, $p_{ij}$ is a probability measure of all points $N$. The initial version of t-SNE is operated for all points, and later improvement used kNN to select neighbor connected points for computing  \cite{van2014accelerating,bohm2022attraction}.
The similarity measure in the low dimension can be defined as 
\begin{equation}
{q_{ij}}={\frac{w_{ij}}{Z}},  {w_{ij}}={\frac{1}{1+d_{ij}^{2}}},  {d_{ij}}={||y_{i}-y_{j}||}, {Z}={\sum_{l\neq r}{w_{lr}}}
\end{equation}
$q_{ij}$ is a probability measure in the low dimension. The weight $w_{ij}$ was initially defined using Gaussian distribution, and it was observed the exponential tail is too "light" and the heavy-tailed $t$ distribution is used instead, where the polynomial tail offers better performance. This is also seen in many other DR methods where the heavy tail is used. The gradients can be derived analytically as

\begin{equation}
\frac {\partial G_{t-SNE}}{\partial y_{i}}=\sum_{j} 4\frac{(p_{ij}-q_{ij})d_{ij}}{(1+d_{ij}^{2})}{e_{ij}}
\end{equation}
where $e_{ij}$ is the unit vector corresponding to $d$ dimension. As can be observed in the above equation, there is a positive component and a negative component. The negative component should be maximized, and that only holds for distant points when the absolute value of $\frac{d_{ij}}{(1+d_{ij}^{2})}$ is minimized. Thus, we can approximate the positive component using only the nearest neighbor and the negative component using distant points. If we use the sign convention as a positive for contraction force, the equation can be rewritten as 

\begin{equation}
\frac {\partial G_{t-SNE}}{\partial y_{i}}={\sum _j{F_{attractive}-F_{repulsive}}}=\sum_{j=1}^{k} 4\frac{(p_{ij})d_{ij}}{(1+d_{ij}^{2})}{e_{ij}}-\sum_{j=k+1}^{N} 4\frac{(q_{ij})d_{ij}}{(1+d_{ij}^{2})}{e_{ij}}
\end{equation}
As can be seen, the gradients of the t-SNE method can be written as a force field with contraction force and repulsive force, which are separated by the k-nearest neighbor.

\subsection{Method: LargeVis}
LargeVis follows a similar concept by separating the data points into two groups: neighbors and distant points. This greatly helps the scalability compared with the original t-SNE formulation. A unique difference of LargeVis is that it uses a loss function compared to the t-SNE, where a free parameter $\gamma$ is used to tune the positive and negative components. The gradients using the LargeVis loss function can be expressed as
\begin{equation}
\frac {\partial G_{LargeVis}}{\partial y_{i}}={\sum _j{F_{attractive}-F_{repulsive}}}=\sum_{j=1}^{k} 4\frac{(p_{ij})d_{ij}}{(1+d_{ij}^{2})}{e_{ij}}-\gamma \sum_{j=k+1}^{N} 4\frac{d_{ij}}{(\varepsilon + d_{ij}^{2})(1+d_{ij}^{2})}{e_{ij}}
\end{equation} 
where $\varepsilon$ is a small number to avoid division of zero. It can be seen that the first positive component (contractive force) is the same as it is in the t-SNE. The repulsive force is different. $\gamma$ can be considered as a weight factor to scale the repulsive force. In LargeVis \cite{tang2016visualizing}, the repulsive force is not computed for every distant edge due to computational cost. Instead, it only samples a certain number of points from distant points to estimate the repulsive force. This is called negative sampling. The equation can be revised as
\begin{equation}
\frac {\partial G_{LargeVis}}{\partial y_{i}}={\sum _j{F_{attractive}-F_{repulsive}}}=\sum_{j=1}^{k} 4\frac{(p_{ij})d_{ij}}{(1+d_{ij}^{2})}{e_{ij}}-\gamma ^{'}\sum_{j}^{m \leftarrow (N-k)} 4\frac{d_{ij}}{(\varepsilon + d_{ij}^{2})(1+d_{ij}^{2})}{e_{ij}}
\end{equation} 
$m$ is the number of negative samples, and they are sampled from the $N-k$ distant points. It should be noted that $\gamma ^{'}$ is used to replace $\gamma$ as the negative sampling will change the weight parameter. However, in many DR methods, the weigh factors need to be tuned anyway, and no attention is paid to the relationship of this tuning parameter with respect to the other model settings, such as the sampling rate.

\subsection{Method: UMAP}
UMAP used the membership strength of two fuzzy sets ( $v_{ij}$ in high dimension and $w_{ij}$ low dimension) to define the similarity measures. This is functionally similar to the probability measure in t-SNE (e.g., $p_{ij}$ and $q_{ij}$). Specifically, UMAP defines 

\begin{equation}
v_{ij}=exp(-\frac{(r_{ij}-Q_{i}}{\sigma _{i}}), {w_{ij}}={\frac{1}{1+ad_{ij}^{2b}}}
\end{equation} 
where $Q_{i}$ is the shortest distances among all neighbor point $i$. For $w_{ij}$, UMAP includes two additional shape factors $a$ and $b$. These two factors mainly control the scale and tail behavior of the function. If $a=1$ and $b=1$, it reproduces the commonly used one in t-SNE and LargeVis. UMAP used a similar loss function setting as in the LargeVis and also employed the negative sampling. Thus, the gradients can be expressed as
\begin{equation}
\frac {\partial G_{UMAP}}{\partial y_{i}}={\sum _j{F_{attractive}-F_{repulsive}}}=\sum_{j=1}^{k} 4\frac{(v_{ij})ad_{ij}^{2b-1}}{(1+ad_{ij}^{2b})}{e_{ij}}-\sum_{j}^{m \leftarrow (N-k)} 4\frac{b(1-v_{ij})d_{ij}}{(\varepsilon + d_{ij}^{2})(1+ad_{ij}^{2b})}{e_{ij}}
\end{equation} 
If $a=1$ and $b=1$, it can be expressed as 

\begin{equation}
\frac {\partial G_{UMAP}}{\partial y_{i}}={\sum _j{F_{attractive}-F_{repulsive}}}=\sum_{j=1}^{k} 4\frac{(v_{ij})d_{ij}}{(1+d_{ij}^{2})}{e_{ij}}-\sum_{j}^{ m \leftarrow (N-k)} 4\frac{(1-v_{ij})d_{ij}}{(\varepsilon + d_{ij}^{2})(1+d_{ij}^{2})}{e_{ij}}
\end{equation} 
It is seen that UMAP and LargeVis have very similar formulations for this special case. This might explain why the two models have very similar performance, although they originated from different settings.

\subsection{Method: PaCMAP}
PaCMAP \cite{wang2021understanding} introduces the concept of mid-near point and includes additional gradient contributions (forces) from these mid-near points. These mid-near points also generate contraction forces. Thus, at the conceptual level, the gradients of PaCMAP can be written as:

\begin{align}
\frac {\partial G_{PaCMAP}}{\partial y_{i}}={\sum _j{F_{attractive, neighbor}+F_{attractive, mid-near}-F_{repulsive}}} \notag\\
=\sum_{j=1}^{k} \frac{w_{NN}}{(11+d_{ij}^{2})^{2}} + \sum_{j=mid-near} \frac{w_{MN}}{(10001+d_{ij}^{2})^{2}} - \sum_{j=far-points} \frac{w_{FP}}{(2+d_{ij}^{2})^{2}}    
\end{align}

where $w_{NN}$, $w_{MN}$, and $w_{FP}$ are the weight factors for near neighbor, mid-near, and far points, respectively. The coefficients in the denominator are suggested by the PaCMAP. Their effects are similar to the scaling and tail factors in UMAP. It is unclear what the mechanism of mid-near points, at least not from the equation. The functionality of mid-near points is very similar to neighbor points, which has been shown that the impact on the final results is minimal for many investigated datasets \cite{jimelvilleNotesonPACMAP}.

\subsection{Method: TRIMAP}
TRIMAP \cite{amid2019trimap} used triplets instead of a pair of points to define gradient computing. For a point $i$, one can define two other points $f$ and $s$. One point is closer to $i$ and is assumed to be point $f$, which generates the contraction force. The other point $s$ will generate the repulsive force for $i$. The gradient can be expressed as

\begin{equation}
\frac {\partial G_{TRIMAP}}{\partial y_{i}}= {\sum _f{F_{attractive, neighbor}-\sum _s F_{repulsive}}}= \sum_{f} \frac{d_{if}}{(3+d_{if}^2+d_{is}^2)^2} - \sum_{s} \frac{d_{is}}{(3+d_{if}^2+d_{is}^2)^2}
\end{equation} 
As the equation shows, the triplets can also be divided into the neighbor (contraction) and distant (repulsion) groups. The force field function also has a polynomial tail, which is used in many other DR methods.

\subsection{Short summary for DR methods}
The above brief review only covers several widely used DR methods, which are very relevant to the proposed CAMEL methods. It should be noted that similar behavior also exists in some other DR methods not reviewed here. For example, it has been shown that the Laplacian Eignemap and ForceAtlas2 methods also included similar attractive and repulsive force groups \cite{bohm2022attraction}. In a recent study \cite{draganov2023unexplainable}, it is shown that the classical PCA and LLE methods can also be viewed in terms of the attractive-repulsive force formulation. Details are not included here as I did not intend to have an exhaustive list of all DR models for the attraction-repulsion discussion. Rather, several common observations are highlighted here:
\begin{enumerate}
    \item Most (if not all) DR models rely on the kNN algorithm to define the neighbor points, which contributes to the contraction force.
    \item It appears that the weight (affinities) in the high dimension is not in constructing the low dimensional embedding if kNN is known. 
    \item Most DR methods use the heavy tail functions (e.g., polynomial tail) to describe the force field as a distance function.
    \item Most DR methods treat the weight coefficients in the force field as tunable parameters without discussing their mechanism concerning model settings.
    \item All DR methods (to the author's best knowledge) only use the pair-wise force field, and the multi-body force field has not been explored.
    \item All DR methods (to the author's best knowledge) only used the distance metrics and other topological descriptors have not been used.
\end{enumerate}
The above summary provides useful insights into new developments. The first three suggested possible characteristics as those have been demonstrated with success from multiple DR models. The latter three are the major motivations of the proposed study to develop a mechanics/physics-guided DR model, CAMEL. Details are shown below.

\section{CAMEL method and algorithm}
The above review clearly shows that the attractive-repulsive force field can be used to explain DR models, which brings up a natural question. Can the centuries of development in mechanics/physics benefit the DR models? This section will start from a unique concept in mechanics/physics, many-body potentials and force, and extend it to the possible DR formulation.

\subsection{Many body potential and force}
As one basic concept in mechanics and physics, force has been extensively investigated for thousands of years. One classical formulation of force is to define it using pairwise (also called local) interactions, such as the classic Hook's law. Another type is the many body (also called non-local) force, which depends on a group of objects that are not necessarily directly linked with each other. This concept has been widely used in molecular dynamics \cite{daw1993embedded,daw1984embedded}, lattice particle method \cite{chen2014generalized,chen2016non}, peridynamics \cite{silling2010peridynamic}. The lattice particle method (LPM) is discussed below to illustrate the concept because it was developed by the same author.

\begin{figure}
  \centering
  \includegraphics[scale=0.5]{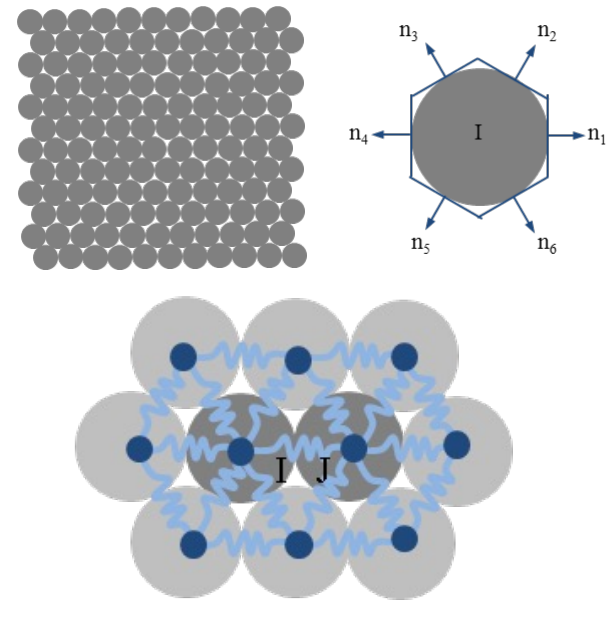}
  \caption{Schematic illustration of neighbor forces of a hexagon lattice unit cell}
  \label{fig:fig1}
\end{figure}

A nonlocal lattice particle model (LPM) was developed aiming to solve some classical mechanics problems using particles located on regular 2D/3D lattice, such as simple cubic, face-center cubic, and body-centered cubic\cite{chen2014generalized,chen2016non}. In LPM, each particle is connected with its neighbors via spring bonds. The spring bonds can be linear-elastic \cite{chen2014generalized,chen2016non} or can be nonlinear \cite{wei2020nonlocal,meng2021modeling}. Each particle interacts with each of its neighbors through pair-wise and many-body potentials. A material particle can have multiple layers of neighbors. Each layer of neighbors has distinct distances from the central particle and can construct a specific unit cell \cite{chen2014generalized}. It has been shown that more neighboring layers tend to smooth out the mechanical anisotropy, such as fracture path \cite{chen2014generalized}. The following discussion uses a single-layer configuration for easy discussion. A schematic illustration for a hexagon packing lattice can be seen in \ref{fig:fig1}.

The potential energy of a material particle consists of the pairwise potential energy of individual bonds and the many-body potential energy of all the neighboring bonds. The nonlocal energy contribution allows LPM to simulate arbitrary Poisson’s ratio in solid deformation problems \cite{chen2014novel}. The potential energy is calculated for a unit cell and the summation of the unit cell will provide the potential energy for the whole system. The total potential energy  can be expressed as

	\begin{align}
    U = {U_{cell}^c} &= ({U_{local}^c}+{U_{nonlocal}^c}) \\
    {U_{local}} &=  \frac{1}{2} K \sum_{j=1}^{k} (\delta {l_{ij}})^2 \\
    {U_{nonlocal}} &= \frac{1}{2} T \left(\sum_{j=1}^{n} \delta {l_{ij}}\right)^2 
	\end{align}

\noindent
where $K$ and $T$ are the model parameters for a bond connecting material particles $i$ and $j$, $\delta l_{ij}$ is the bond elongation, and $k$ is the number of bonds for the  particle $i$. Note that I used the same $k$ as the number of $k$ nearest neighbors in the DR methods. $K$ represents the pairwise stiffness of the spring bond between particle $i$ and its surrounding neighbors \cite{chen2014novel}, while $T$ represents the many-body stiffness of all neighbors for particle $i$ and $j$. The interaction force within the bond $ij$ can be obtained by differentiating the total potential energy of the particle $i$ with respect to the bond elongation. The absolute value of the force exerted from particle $j$ to particle $i$ is $f_{ij}=\frac{\partial U^i}{\partial \left( \frac{1}{2} \delta l_{ij}\right)}$ \cite{chen2014novel}. By taking an average of $f_{ij}$ and $f_{ji}$ (similar techniques used in t-SNE and others for symmetry), the total bond force $\boldsymbol{F}_{ij}$ between particle $i$ and $j$ is
\begin{equation}
	\boldsymbol{F}_{ij} = \frac{1}{2}(f_{ij}{\boldsymbol{e}}^{ij}-f_{ji}{\boldsymbol{e}}^{ji})=\left( K (\delta l_{ij}^c+\delta l_{ji}) + T\left( \sum_{l=1}^{k} \delta l_{il} + \sum_{s=1}^{k} \delta l_{js} \right) \right) {\boldsymbol{e}}^{ij}
\end{equation}

\noindent
where ${\mathbf{e}}^{ij}$ (and ${\mathbf{e}}^{ji}$) is the unit bond vector between material particles $i$ and $j$, which is given for a specific lattice geometry. This unit vector is also used in the above discussion for DR methods and uses the same notation here. It should be noted that, due to the nonlocal nature, the neighbors of particle $j$ will also contribute to the forces of particle $i$. As can be seen, the forces field of the particle system is an analogy to the contraction force in the DR methods discussed above. In the LPM \cite{chen2014novel}, the many-body potential is defined as the volumetric change of each particle neighbor, which is a density measure. The main reason is to explain the dilatational and distortional deformation and Poisson's ratio effect in solids \cite{meng2021modeling,wei2020nonlocal,chen2014novel}. For the DR problems, if the many-body force is included in the contraction force computing, the detailed many-body force needs to be determined, inspired by the following topology discussion.     

\subsection{Curvature in topology}
\begin{figure}
  \centering
  \includegraphics[scale=0.5]{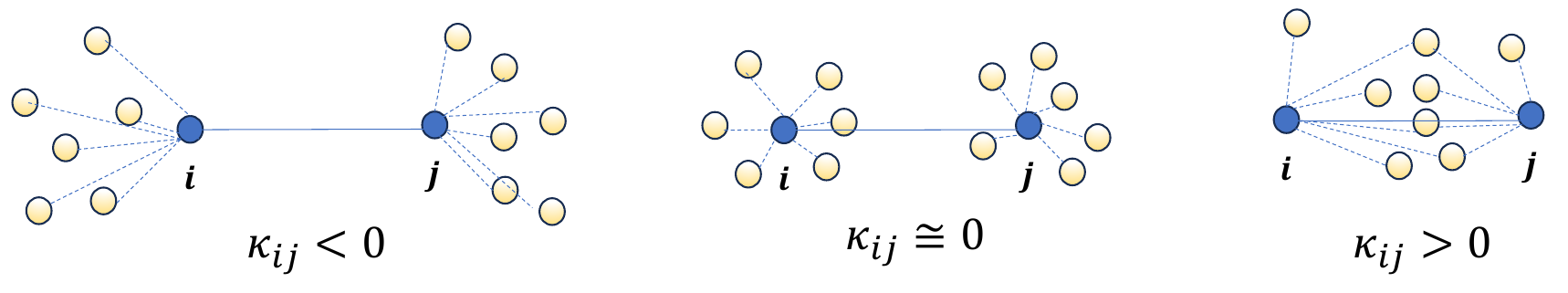}
  \caption{Schematic illustration of different curvature for a pair of points $i$ and $j$}
  \label{fig:fig2}
\end{figure}

Many DR methods are based on the kNN graph, which can be considered as a topological descriptor of the high-dimensional data using the distance metrics (e.g., Euclidean distance for most DR methods). In addition to the distance, the curvature of another descriptor for the topology. In Riemannian geometry, curvature defines how a shape deviates from Euclidean. For example, Ricci curvature is used in continuous space \cite{bochner1946vector} and Ollivier-Ricci curvature \cite{ollivier2009ricci}, and its variation \cite{lin2011ricci} has been used to describe the discrete graph curvature. The Ollivier-Ricci curvature for a pair of node $i$ and $j$ is defined as 

\begin{equation}
    \kappa _{ij}=1- \frac{W(m_i,m_j)}{d_{ij}}
\end{equation}
where $\kappa _{ij}$ is the Ollivier-Ricci curvature of edge $ij$. $d_ij$ is the direct distance of edge $ij$. $W(m_i,m_j)$ is the Wasserstein distance (or Earth Mover distance) which finds the optimal transportation plan between two probability measures $m_i$ and $m_j$. The probability is defined by neighbor points of $i$ and $j$, e.g., uniform, Gaussian-like, or triangular distribution. The curvature is positive if the Wasserstein distance is less than the direct distance. In General Relativity, positive curvature will cause gravity and attraction, whereas negative curvature will cause repulsion. The intent here is not to use the General Relativity and define the force field rigorously. Rather, the concept of curvature-induced force is used here.  An illustration is shown in \ref{fig:fig2} for three different curvature cases. When the neighbor points of point $i$ and $j$ are far away in the graph, the Wasserstein distance is usually larger as the traveling distance in the graph network will be larger. Thus, a negative curvature is obtained. This negative curvature feels that a repulsive force is pushing the points away from each other.  If the neighbor points are close to the points $i$ and $j$, the Wasserstein distance is approximately equal to the direct distance between $i$ and $j$. The curvature is zero, and the geometry is almost "flat". There is no attraction and repulsion between points $i$ and $j$. When the neighboring points are close to the center of the points $i$ and $j$, many of them are common neighbors. Thus, the Wasserstein distance is less than the direct distance between $i$ and $j$, and the positive curvature is obtained. In extreme cases, the points $i$ and $j$ share the same neighbors, and the Wasserstein distance is zero (no mass needs to be moved in the graph). The curvature of unifying is obtained. The positive curvature feels that the points $i$ and $j$ are attracted to be together by a force. This intuition inspired the author to revise the force field in the DR methods by including a curvature-induced force term. The curvature of the graph has been successfully used to identify the community of the network in \cite{ni2019community}. No explicit force field is defined, and only the attraction-repulsion is used to identify the community in a network. 

The original Ollivier-Ricci curvature definition in the graph must compute the Wasserstein distance in a network, which involves the shortest distance search. This is computationally prohibitive for large networks (e.g., kNN from large-scale datasets). Thus, an alternative definition of curvature, CAMEL-Curvature, is proposed. From Figure \ref{fig:fig2}, the Ollivier-Ricci curvature is closely correlated with the geometric centroid of the neighboring points. Thus, a simple approximation is proposed as

\begin{align}
    \kappa _{ij}=1- \frac{d(c_i,c_j)}{d_{ij}} \notag\\
    c_i = \frac {\sum _{l=1}^k (x_l)}{k},c_j = \frac {\sum _{s=1}^k (x_s)}{k} \notag\\
    d(c_i,c_j)=||c_i-c_j||
\end{align}
where $c_i$ and $c_j$ are the centroid coordinates of point $i$ and $j$, respectively. $d(c_i,c_j)$ is a distance measure (Euclidean distance in this case) of the two centroid. It is seen that this expression matches several special cases in the Ollivier-Ricci curvature computing and is assumed to reflect a curvature measure of point $i$ and $j$. The benefit of this definition is that it only involves simple arithmetic operations and is very fast to implement.

\subsection{Proposed force field model}
Based on the above discussion, a new force model is proposed with the following hypothesis.
\begin{enumerate}
    \item In DR computing, each data point is a "particle" and feels that forces from other particles (data points)
    \item The force field can be decomposed into three components: an attractive force by neighbor points, a repulsive force by the distant points, and a force by the curvature of neighbor points (can be attractive or repulsive depending on the curvature)
    \item attractive force and repulsive force can be described as a heavy-tailed Pareto \cite{arnold2014pareto} distance function 
\end{enumerate}

\begin{figure}
  \centering
  \includegraphics[scale=0.5]{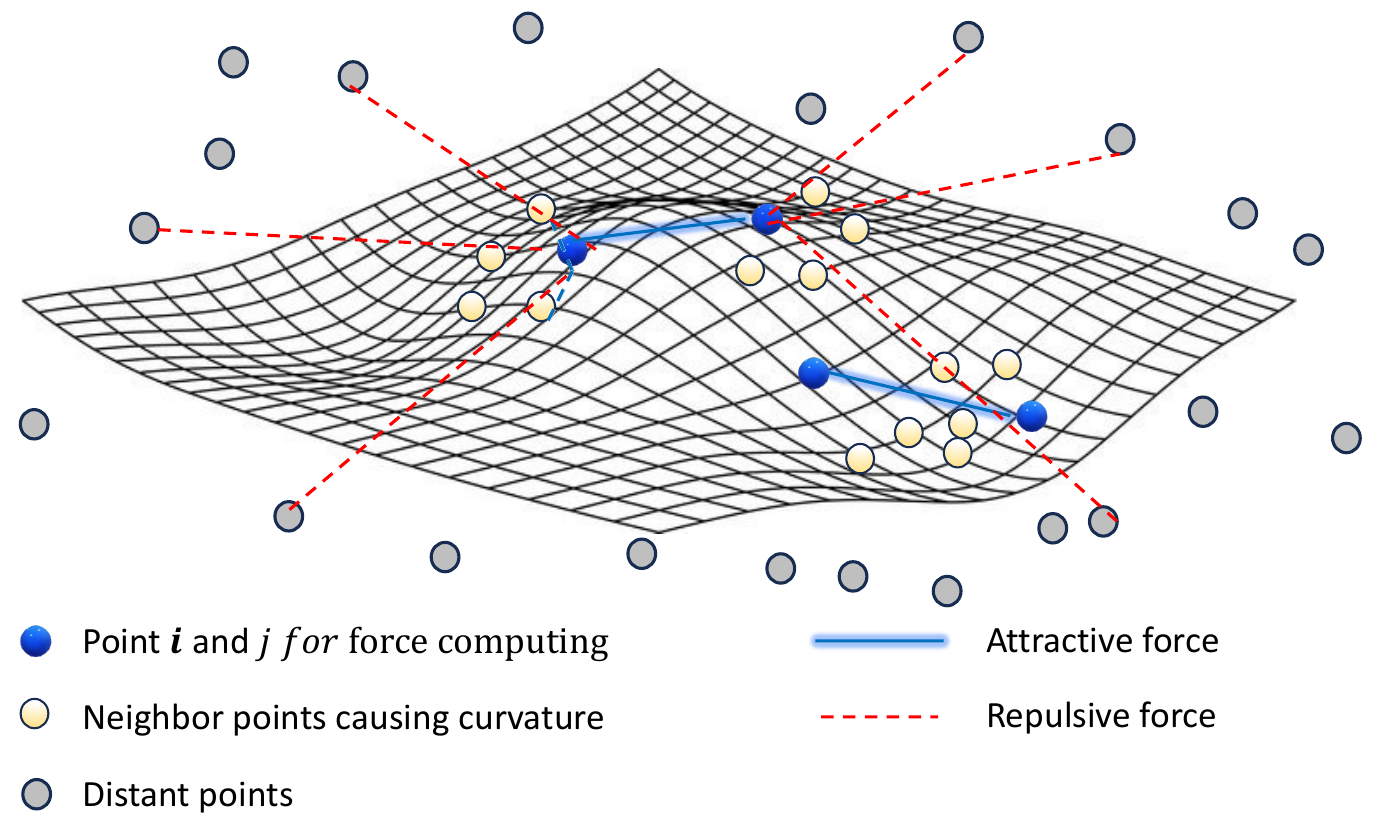}
  \caption{Schematic illustration of force field model for pair of points $i$ and $j$}
  \label{fig:fig3}
\end{figure}

The schematic illustration of the proposed model is shown in Figure \ref{fig:fig3}. The blue points are the points to be considered for force computing. The yellow points are the neighbor points, which will cause the curvature-induced force. As can be seen, different signs of curvatures exist and will change the force to either attractive or repulsive. Grey points are distant points, which will cause the repulsive force. It is shown that only a portion of the distant points is used for computing the repulsive force, which is known as negative sampling. Following the same notation in the method review section, the proposed force field for CAMEL can be expressed 
\begin{align}
\frac {\partial G_{CAMEL}}{\partial y_{i}}=\sum_{j=1}^k {F_{attractive, neighbor}+\sum_{j=1}^k F_{curvature}- \sum _{j=k+1}^{N} F_{repulsive}} \notag\\
=\sum_{j=1}^{k} w_{NN}\frac{1}{(1+ \frac {d_{ij}^{2}}{\sigma})^{2}} + \sum_{j=1}^{k} w_{CR}(1- \frac {d(c_i,c_j)}{d_{ij}}) - \sum_{j=k+1}^N w_{DP}\frac{1}{(1+d_{ij}^{2})^{2}}        
\end{align}
where $w_{NN}$, $w_{DP}$, and $w_{CR}$ are the weight coefficients for Near Neighbor (NN), Distant Points (DP), and CuRvature (CR), respectively. The last term for DP is proportional to the number of data points $N$, for which computing is expensive for large datasets. Thus, many DR algorithms only choose a subset of $N$ to compute the repulsive force, which is called "negative sampling." If only $m$ DP points are selected for force computing, the above equation can be rewritten as 
\begin{align}
\frac {\partial G_{CAMEL}}{\partial y_{i}}=\sum_{j=1}^k {F_{attractive, neighbor}+\sum_{j=1}^k F_{curvature}- \sum _{j=k+1}^{N} F_{repulsive}} \notag\\
=\sum_{j=1}^{k} w_{NN}\frac{1}{(1+ \frac {d_{ij}^{2}}{\sigma})^{2}} + \sum_{j=1}^{k} w_{CR}(1- \frac {d(c_i,c_j)}{d_{ij}}) - \sum_{j=1}^m w_{DP}\frac{1}{(1+d_{ij}^{2})^{2}}        
\end{align}

This proposed field used the heavy-tailed Pareto distribution for the square distance function. The probability density function of the Pareto distribution can be expressed as 

\begin{align}
f_x(x)=(\frac{1}{1+\frac{x-\mu}{\sigma}} )^{\alpha}
\end{align}
where $\mu$ is the location parameter and is zero in this case. $\alpha$ is the tail index and is assumed to be 2 following the suggestions in TRIMAP \cite{amid2019trimap} and PaCMAP \cite{wang2021understanding}. $\sigma$ is the scale parameter and is assumed to be 20 for the near neighbor and 1 for the distant points. This difference ensures that the contraction force among neighbor points decreases slower than the repulsive force so they can be grouped together.

Next, let us see the impact of the number of neighbors and the number of negative sampling points on the coefficients. Let us consider the extreme case when the equilibrium is achieved at the end of particle dynamics (e.g., the gradients are zero). We use the average near-neighbor distance, average Distant Points distance, and average curvature to rewrite the force equation above as
\begin{align}
0=k*w_{NN} \frac {1}{(1+\frac{\bar{d_{ij}}^2}{\sigma})^2}+k*w_{CR}(1- \frac {\bar{d(c_i,c_j)}}{\bar{d_{ij}}})-(N-k) w_{DP} \frac {1}{(1+\bar{d_{ij}}^2)^2}    
\end{align}
where all $\bar{d}$ indicates the average value. If the negative sampling is used, only $m$ distant points are used to compute the repulsive force. Thus, the equation can be expressed as 

\begin{align}
0=w_{NN} \frac {\bar{d_{ij}}}{(1+\frac{\bar{d_{ij}}^2}{\sigma})^2}+ w_{CR}(1- \frac {\bar{d(c_i,c_j)}}{\bar{d_{ij}}})-w_{DP}^{'} \frac {\bar{d_{ij}}}{(1+\bar{d_{ij}}^2)^2}
\notag\\
w_{DP}^{'}=\frac{m}{k} w_{DP}
\end{align}       
where $w_{DP}^{'}$ is the weight factor for many DR algorithms for the repulsive force. As can be observed, $w_{DP}^{'}$ depends on the intrinsic weight factor $w_{DP}$ (per edge for distant point) and model settings for the number of neighbor points in the kNN graph and negative sampling ratio. The amplification factor is $\frac{m}{k}$. This result makes sense from a mechanistic point of view. If more distant points are used to supply the same amount of repulsive force at the equilibrium, each distant point will provide less force. Since the distance of distant points to the current point is statistically equal due to random sampling, the linear proportion is expected for the scale factor. This explains why many DR models are sensitive to the number of neighbors and negative sampling ratio and need to calibrate the weight factors for different settings. In the later discussion section, it is demonstrated that the derived weight amplification factor stabilized the DR performance irrespective of the model settings, where the performance is not stable when the amplification factor is ignored.

The above discussion only uses the kNN graph from the high dimension and assumes the uniform weight for all neighbors (i.e., only use the adjacency matrix without considering the distance difference among neighbors). Also, the curvature uses the absolute curvature computed from the low-dimensional space only and does not consider the relative change from the high-dimensional space. In order to include the high-dimensional weight in the embedding computing, modification of the force weight coefficient is proposed as 
\begin{align}
\frac {\partial G_{CAMEL}}{\partial y_{i}}=\sum_{j=1}^k {F_{attractive, neighbor}+\sum_{j=1}^k F_{curvature}- \sum _{j=k+1}^{N} F_{repulsive}} \notag\\
=\sum_{j=1}^{k} w_{NN}\frac{1}{(1+ \frac {d_{ij}^{2}}{\sigma})^{2}} + \sum_{j=1}^{k} w_{CR}(1- \frac {d(c_i,c_j)}{d_{ij}}) - \sum_{j=1}^m w_{DP}\frac{1}{(1+d_{ij}^{2})^{2}} \notag\\
w_{NN}=[1-arctan(\frac{D_{ij}}{\hat{D_{ij}}}-1)\frac{1}{\pi}]w_{NN}^{'}\notag\\
w_{CR}=\frac{(C_ij-(1- \frac {d(c_i,c_j)}{d_{ij}}))}{(1- \frac {d(c_i,c_j)}{d_{ij}})}w_{CR}^{'}\notag\\
w_{DP}=[1+arctan(\frac{D_{ij}^{DP}}{\hat{D_{ij}^{DP}}}-1)\frac{1}{\pi}]\frac{k}{m}w_{DP}^{'}\notag\\
\label{Eq: final force field}
\end{align}

where ${D_{ij}}$ and ${\hat{D_{ij}}}$ are the pair-wise distance of the neighbor and the average distance of all neighbor points, respectively. The $arctan()$ function enforces the symmetric bounded distribution of the scaling factor in the $[0.5,1.5]$ range with a mean of 1.0. For the closer neighbor points, the weight coefficient approaches 1.5. Similarly, $D_{ij}^{DP}$ and ${\hat{D_{ij}^{DP}}}$ are the pair-wise distance of the distant points and the average distance for all distant points, respectively. The difference is that the distance points farther away have a weight coefficient approaching 1.5, as they should have a larger repulsive force during the embedding optimization. $C_ij$ is the curvature computed from the high dimensional data. Thus, the curvature-induced force will approach zero if the curvature difference between the high-dimension and low-dimension is zero. This is the mechanism by which the low-dimension embedding will search for the arrangement that minimizes this difference. The later section will show the effect of this term on the curvature similarity metric. As a summary, Eq. \ref{Eq: final force field} is the final force field of the proposed CAMEL.

\section{Algorithm and Implementation}
The above discussion is for the theoretical foundation of the proposed CAMEL. This section focuses on its algorithmic implementation with settings. The focus is to develop CAMEL algorithms suitable for common machine learning tasks. It should be noted that most existing DR methods and codes aim for unsupervised learning, which is one of the main focuses of DR methods. Few studies (e.g., UMAP) have also developed other algorithms for supervised, metric, semi-supervised, and inverse generative learning. Thus, one of the primary motivations and objectives of the current study is to provide a systematic and rigorous formulation of different machine learning tasks in the CAMEL framework. In addition, verification and validation of different DR methods are usually performed with visual comparison with limited evaluation metrics. Another goal in this section is to provide and/or develop a holistic list of evaluation metrics for comparison. Interested readers can also use the listed metrics to evaluate similar approaches. Details are shown below. 

\subsection{Unsupervised learning}
Unsupervised learning is formulated in a similar way as most DR methods: giving input data $X$ with dimension as $[n_samples, n_features]$, an unsupervised DR wants to find an embedding $Y$ with dimension as $[n_samples, n_components]$. The $n_components$ is the desired lower dimension, e.g., 2 in most examples in this paper, but it can be an arbitrary number, even a high number for inverse problems of generative modeling. No label information is given. A pseudo-code for the algorithm is shown below.The algorithm is developed in Python with numba to accelerate the computing. The framework follows the TRIMAP and PaCMAP templates. The significant difference is in the curvature computing and force calculation (i.e., gradients calculation). Since this approach formulation is well-documented in most DR methods and is similar, no specific discussions are provided here.

\begin{algorithm}
	\caption{CAMEL-unsupervised learning}
	\begin{algorithmic}[1]
		\State Input data points $X_i$ and preprocessing (centering and scaling)
  		\State kNN graph construction using ANNOY (https://github.com/spotify/annoy) algorithm
    	\State Generate samples for neighbor and distant points (negative sampling)
     	\State Embedding initialization using PCA or random
            \For {iteration < max iter number}
		\State Compute the curvature using the current embedding
  		\State Compute the attractive and repulsive forces
		\State ADAM optimizer for embedding optimization
		\EndFor
		\State Output embedding
	\end{algorithmic} 
\end{algorithm} 

\subsection{Supervised learning}
Not many DR methods discuss the supervised learning setting. Some treat supervised learning as an unsupervised learning problem with augmented feature space using label information. This approach alters the original feature space structure. It makes semi-supervised learning/metric learning difficult, as the label information is usually not available for new unobserved data or is missing for some training data. It is also not ideal for an inverse generative model where the feature structure rather than label information is more important. Thus, the proposed model uses a different formulation. The semi-supervised learning in CAMEL is formulated as giving input data $X$ with dimension as $[n_samples, n_features]$ and $label$ with dimension as $[n_samples, n_class]$, a supervised DR wants to find an embedding $Y$ with dimension as $[n_samples, n_components]$. $n_class$ is the number of classes in the label, which could be categorical (for classification) or numerical (for regression). It can also be a multi-class label with mixed categories. 
The key element in the proposed CAMEL (and many other DR methods) is the knn graph. Thus, the label information is included in the DR by revising the original knn graph using the label information. Once the new knn graph is available, the label information can be removed and the original feature space with the revised knn graph can be used for DR following the same procedure as in the unsupervised learning case. Since the knn graph is constructed by using a "distance" measure, the label information needs to be transformed for distance measure. This is straightforward for numerical labels. CAMEL used the "one-hot encoder" for transformation to ensure the distances between categories are equal for categorical labels. Another treatment is the scaling of label distance. Since the feature distance is from high-dimensional data and can be arbitrarily large, the inclusion of the label distance needs to be weighted to be appropriately included in the knn graph construction. The proposed CAMEL uses the following scaling, i.e.,
\begin{align}
label^{'}=\frac {w_label}{1-w_label}*\frac{\bar{dist_X}}{\bar{dist_label}}*label
\end{align} 
where $\bar{dist_X}$ and $\bar{dist_labe}$ are the average Euclidean distance of feature $X$ and label $label$, respectively. Since only approximate ratios are needed here for scaling, a subset of all data points are sampled to estimate this ratio. $w_label$ is a model weight parameter that reflects how important the label information is when revising the knn graph. If $w_label$ is zero, the label distance is zero in the knn computing. Thus, no label information is used to revise the knn graph. If $w_label$ is unity, the label distance is infinity in the knn computing. Thus, only label information is used to revise the known graph. The default value of $w_label$ is 0.5, and both label and feature information are used to revise the knn with an approximate ratio of 1:1. In the later discussion section, examples with different $w_label$ factors are used to illustrate its impact on the final embedding. A pseudo-code is provided below. 
\begin{algorithm}
	\caption{CAMEL-supervised learning}
	\begin{algorithmic}[1]
		\State Input data points $X_i$ and preprocessing (centering and scaling)
  		\State Input label $label$ and transform to numerical (one-hot encoding for categorical)
      	\State Scale feature vector and label vector and concatenate
  		\State kNN graph construction using ANNOY (https://github.com/spotify/annoy) algorithm
      	\State Remove label from the concatenated array
    	\State Generate samples for neighbor and distant points (negative sampling)
     	\State Embedding initialization using PCA or random
            \For {iteration < max iter number}
		\State Compute the curvature using the current embedding
  		\State Compute the attractive and repulsive forces
		\State ADAM optimizer for embedding optimization
		\EndFor
		\State Output embedding
	\end{algorithmic} 
\end{algorithm} 

\subsection{Metric learning}
Metric learning refers to learning a metric that can be used to map unseen new data to the embedding space. Many DR methods do not discuss the metric learning setting and do not differentiate training data and new data embedding. For example, the original t-SNE model treats the new data the same way as the original data and performs the new embedding with all data together. A similar approach has been used in PaCMAP, and the new data is concatenated with the training data to do the embedding again. One direct consequence of this type of approach is that the embedding will change as the new data is processed. This setting is counter-intuitive with the metric learning or testing data. The training data is used to learn a model, and the model is used to map the new data (testing) to the embedding space. Thus, the model should not be changed with the new testing data; otherwise, the data "leakage" happens. Thus, the proposed model treats metric learning as a training/testing problem and differentiates the training and testing data. The metric learning in CAMEL is formulated as giving training data $X$ with dimension as $(n_{samples}, n_{features})$ and testing data $X^{'}$ with dimension as $(n_{samples}^{'}, n_{features}^{'})$, a metric learning DR wants to find an embedding $Y^{'}$ with dimension as $(n_{samples}^{'}, n_{components}^{'})$ with already solved embedding $Y$ of training data with dimension as $(n_{samples}, n_{components})$. In metric learning, $label$ may or may not be available. The key step in metric learning is to "freeze" the embedding of the training data during the testing data embedding. The knn of the new testing data is constructed with respect to the original training data knn. Thus, the attractive/repulsive force field from the learned embedding space will automatically map the new testing data in the original embedding space.  

\begin{algorithm}
	\caption{CAMEL-metric learning}
	\begin{algorithmic}[1]
            \State Input training points $X_i$ (with/without $label$) and perform CAMEL for initial embedding $Y_i$		
            \State Input data points $X_i^{'}$ and preprocessing (centering and scaling) using the original training data parameters
  		\State kNN graph for new testing data using ANNOY algorithm and initial embedding $Y_i$	
    	\State Generate samples for neighbor and distant points (negative sampling)
     	\State Embedding initialization using PCA or random
            \For {iteration < max iter number}
		\State Compute the curvature using the current embedding
  		\State Compute the attractive and repulsive forces
		\State ADAM optimizer for embedding optimization
		\EndFor
		\State Output embedding for new testing data $Y_i^{'}$	
	\end{algorithmic} 
\end{algorithm} 

\subsection{Semi-supervised learning}
Semi-supervised learning refers to the case that some data's labels are missing and some data's labels are available. One straightforward way for semi-supervised learning is to drop the data with missing labels during the training phase and use the learned embedding to perform metric learning for unlabeled data. Thus, combined supervised learning and metric learning are used for semi-supervised learning. This approach, however, does not use the structure information of the unlabeled data for the embedding and will have poor performance when the available labeled data is limited, and the leaned embedding is of poor quality. Thus, the proposed CAMEL will use an alternative approach, which is data imputation. The semi-supervised learning in CAMEL is formulated as giving data $X$ with dimension as $(n_{samples}, n_{features})$, within which $(n1_{samples})$ data has labels and $(n2_{samples})$ data. Without loss generality, let us assume the first $(n1_{samples})$ data are labeled data with $label$. A semi-supervised learning DR wants to find an embedding $Y$ for all labeled and unlabeled data with dimension as $(n_{samples}^{'}, n_{components}^{'})$.

Since the CAMEL uses the knn as the key element, knn imputation provides estimated labels for the unlabeled data. The average label value from all neighbors is used as the estimated label for numerical labels. For categorical labels, the label with the maximum probability in the neighbor points is used as the estimated label. The knn imputation in the current investigation shows great performance when the percentage of labeled data is relatively high (see details in later discussion sessions). When the percentage of labeled data is low (e.g., less than $5\%~10\%$), the error of knn will increase, and the credibility of knn imputation drops. In the extreme case that one only has one labeled data, the imputation should not be trusted as the imputation results will be random depending on the chosen labeled data. In order to have a smooth transition of unsupervised, semi-supervised, and supervised learning, an imputation credibility function is proposed to revise the label weight during the knn construction as

\begin{align}
label^{'}=\frac {w_{label}^{'}}{1-w_{label}^{'}}*\frac{\bar{dist_X}}{\bar{dist_label}}*label\notag \\
w_{label}^{'}=w_{credibility}*w_{label}=(\frac{1}{2}+\frac{arctan(100*(label_{ratio}-5\%))}{\pi})*w_{label}
\end{align} 
When the label ratio is very low, the credibility function approaches zero, and the semi-supervised learning reduces to unsupervised learning. When the label is large, the credibility function approaches unity. The semi-supervised learning becomes the supervised learning (with estimated labels). The mean of the credibility function is empirically assumed to be $5\%$ based on the collected data in this study. A pseudo-code is provided below. 
\begin{algorithm}
	\caption{CAMEL-semi-supervised learning}
	\begin{algorithmic}[1]
            \State Input training points $X_i$ (with partial labels $label_1$ for the first $n_1$ data)
            \State Construct an initial knn graph using $n_1$ data and impute labels $label_2$ for the remaining $n_2$ data 
            \State Concatenate feature vector $X_{n_1}$ and $X_{n_2}$ and label vector $label_1$ and $label_2$
  		\State Construct a new kNN graph for all data using ANNOY algorithm 
    	\State Generate samples for neighbor and distant points (negative sampling)
     	\State Embedding initialization using PCA or random
            \For {iteration < max iter number}
		\State Compute the curvature using the current embedding
  		\State Compute the attractive and repulsive forces
		\State ADAM optimizer for embedding optimization
		\EndFor
		\State Output embedding for all testing data $Y_i$	
	\end{algorithmic} 
\end{algorithm} 

\subsection{Inverse Learning}
The proposed CAMEL is intrinsically cable of both forward and inverse dimension reduction operations. In the original derivation, there is no restriction on the dimension of the force field. For the forward analysis (dimension reduction), the force field is defined in the lower dimensional space. For the inverse analysis (dimension augmentation), the force field is defined in the higher dimensional space. This is analogous to the well-known encoder-decoder neural network model. The forward dimension reduction is similar to the encoder network and projects the original feature space to a lower dimension latent space. The inverse dimension augmentation is similar to the decoder network and projects the latent space point back to the original feature space.

The inverse learning in CAMEL is formulated as giving data $X$ with dimension as $(n_{samples}, n_{features})$, and an embedding $Y$ is obtained using the CAMEL method. This embedding can be obtained with or without label information. For some newly provided points $Y_{new}$ in the lower dimensional space, find the corresponding $X_{new}$ points in the original feature space. The key concept in inverse learning is that the feature and embedding space are switched during the inverse operation. The knn graph is re-constructed in the lower dimensional space. The force field in the high-dimensional space is assumed to take the same form as that in the lower-dimensional space. Thus, the knn graph in the lower-dimensional space will guide the movement of points in the high-dimensional space to find their equilibrium position. Another difference in inverse learning is the initialization of the projection. Inverse learning does not have PCA initialization, unlike forward analysis, which has PCA and random initialization. An "interpolation-based" initiation is proposed, which takes the initial project as the mean of all neighbor points. The later section will show that both "random" and "interpolation" initialization will generate good inverse learning results. The "interpolation" has faster convergence and less noisy reconstruction. Thus, the CAMEL uses the "interpolation" as the default initialization method. A pseudo-code for inverse learning is provided below.     
\begin{algorithm}
	\caption{CAMEL-inverse learning}
	\begin{algorithmic}[1]
            \State Input training points $X$ (with or without labels $label$ 
            \State Perform the CAMEL embedding (unsupervised, semi-supervised, or supervised) to find embedding $Y$
            \State Construct a new kNN graph using $Y$ using ANNOY algorithm 
    	\State For some points $Y_{new}$, generate samples for neighbor and distant points (negative sampling)
     	\State Inverse embedding initialization using interpolation or random
            \For {iteration < max iter number}
		\State Compute the curvature using the current embedding
  		\State Compute the attractive and repulsive forces
		\State ADAM optimizer for embedding optimization
		\EndFor
		\State Output embedding $X_{new}$ for all testing data $Y_{new}$	
	\end{algorithmic} 
\end{algorithm} 

\subsection{Metrics-based comparison and verification}
A critical element in the proposed study is evaluating the proposed CAMEL and other DR methods. Compared to the well-established evaluation metrics for machine learning tasks, such as classification, regression, generation, etc., the metrics for DR are not very mature. There are no universally accepted metrics for DR evaluation, and many existing studies highly depend on the visual inspection of the final embedding results. Visual inspection is, of course, the first element in evaluation, as many DR methods aim to achieve low-dimensional visualization. However, visual inspection cannot provide quantitative and subjective evaluation. Thus, relatively comprehensive evaluation metrics are used in this study. They are from several recent similar studies, and several metrics are proposed in this study. The main idea is not to find a "ground truth" for the best model or metric. Rather, it provides a complete comparative study to show the differences and similarities when using these metrics to interpret the DR results. 
Earlier studies on the systematic evaluation of DR methods can be tracked back to \cite{van2009dimensionality}, where the focus is on capability comparison. Generalization error is used as the metric for evaluation of the performance. A recent study used the Co-ranking LCMC curves for DR method performance evaluation \cite{fanaee2019performance}. Co-ranking-based metrics \cite{lee2009quality} are widely used for performance evaluation, which is also used in the current study. It should be noted that most DR methods use some "distance measure" in their formulation. The absolute value of these distance measures is difficult to be directly compared between high- and low-dimension due to the scaling issue. Thus, the co-ranking matrix used the ranks at high- and low- -dimensions to form a 2D array to indicate the similarities of different dimensions. In this case, no information for label information is needed, and the evaluation is purely based on $X$ in the feature space and $Y$ in the embedding space. Once the co-ranking matrix $Q$ is computed, many derived metrics using the $Q$ can be developed. For example, trustworthiness and continuity are proposed to use the "intrusion" and "extrusion" of the $Q$ matrix off-diagonal terms as evaluation criteria. The "intrusion" and extrusion" refer to the positive and negative rank error, where high- and low-dimension ranks differ. Local Continuity Meta Criterion (LCMC) is also proposed to use the overlap in a local neighborhood to indicate how good the ranks are consistent in the local $k$ neighborhood. As indicated in \cite{lee2009quality}, the LCMC quantifies the true positives. The trustworthiness and continuity are for the false positives and false negatives, respectively. Another metric using the area under the curve (AUC) of the neighborhood preservation for all possible neighborhood numbers is used as an overall performance metric \cite{lee2015multi}. The AUC covers local and global (smaller $k$ neighborhood numbers) and global (larger $k$ neighborhood numbers). Thus, trustworthiness(Trust), continuity (Conti), LCMC, and AUC are used for quantitative evaluation in the co-ranking matrix-based metrics.

In addition to the co-ranking-based metrics, many other metrics have been used, for example, in \cite{huang2022towards}. Huang et. al. \cite{huang2022towards} proposed three local- and four global-focused metrics for the DR evaluation. support vector machine (SVM-classify) and kNN-based classifiers (kNN-classify) are used with DR embedding to evaluate the local structure preservation capability. These two metrics require the label information and assume that DR methods are for classification jobs. Another local metric for neighbor point preservation (NPP) is also proposed, where the neighbor point intersection of the high-dimensional and low-dimensional neighbor list is used to measure local structure preservation. For global structure preservation, two unsupervised metrics, random triplet (Triplet) and Spearman correlation (SpearCorr), are used. The Triplet metrics evaluate the order preservation of random triple points between the high- and low- dimensions. The SpearCorr used a similar concept by calculating the correlation of rank orders of random pairs. Two supervised metrics are also used for global structure evaluation, which are both based on the centroid rather than individual data points. The centroid kNN (Cen-kNN) class preservation evaluates the fraction remaining in the low-dimensional embedding. The centroid distance (Cen-dist) correlation uses the rank order Spearman correlation for centroids (rather than points).

\begin{figure}
  \centering
  \includegraphics[scale=0.5]{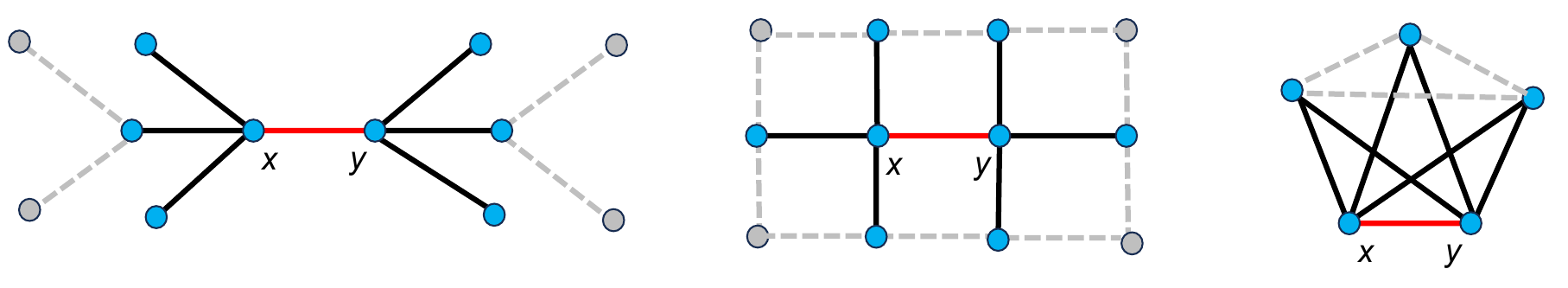}
  \caption{Schematic illustration of same edge distance with different curvature in a graph}
  \label{fig:fig curvature distance}
\end{figure}

The above-mentioned metrics already cover many different perspectives, yet the author still thinks that a few aspects have not been investigated with the above-mentioned metrics. Thus, a few additional metrics are proposed. First, the key concept in CAMEL is to use the curvature information in the model development. Thus, the curvature preservation capability is considered one important addition. Most of the above metrics are based on distance measures. It has been shown that the distance measures are insufficient to describe the data structure's topology \cite{ye2019curvature}. A schematic illustration is shown in \ref{fig:fig curvature distance}. All neighbor edges of $x$ and $y$ are the same distance (e.g., weight 1 for all edges). Their Ollivier-Ricci curvature are negative, zero, and positive, respectively. Thus, the curvature in the high- and low- dimensional space should be similar to preserve this topological pattern. A Curvature Similarity (Curv-Simi) score is proposed by comparing the average curvature of neighbor points in the high- and low- dimension as 

\begin{align}
Score_{Curv-Simi}= \frac{1}{\exp^{\left\| C_{hd} - C_{ld} \right\|}}
\end{align} 
where $C_{hd}$ and $C_{ld}$ are the average curvature in the high dimensional space and low dimensional space, respectively. $\left\| \right\|$ takes the absolute value of the difference. As can be seen, when the difference is zero, the score is 1. If there is a difference, the score will be smaller than 1. For a very large difference, the score is approaching zero. The Curv-Simi score is for local structure evaluation, focusing on the pattern rather than distance.

Another metric is proposed to evaluate the preservation of neighbor points. Since most DR methods use random initiation and optimization, the low-dimensional embedding will have some uncertainties. A few neighbor points differences usually do not change the topological relationship much. Thus, the most critical points are the ones with a large percentage of wrong neighbors, which usually indicates that the topological structure is completely changed between high and low-dimensional space. One intuitive example is the Swiss roll data (see details in later sections), and DR models may not be able to unfold it completely. Some portion of the 3D shape will be overlayed in the 2D projection. In this case, the overlay region neighbor list in the low dimensional will have a higher percentage difference than the high-dimensional neighbor list. Based on this intuition, a metric called  neighbor-not-wrong ratio (NNWR) is proposed as 
\begin{align}
NNWR= 1-\frac{n_{wrong-neighbor}}{n_{data}}\notag  \\
n_{wrong-neighbor}: P(w)>0.5
\end{align} 
where NNWR is a ratio of neighbor-not-wrong points to the total data points. The wrong neighbor point is the point with at least  $50\%$ different neighbor points between high and low dimensions (e.g., the probability of wrong neighbor points $P(w)>0.5$. This metric can be considered a filtered neighbor preservation measure, focusing on the more severe discrepancies rather than small ones. Finally, a last metric for global structure evaluation is proposed. The DR algorithm has a well-known "fragmentation" behavior, which means that low-dimensional embedding will have many small isolated clusters. This is usually caused by the improper balance of the attractive and repulsive forces \cite{bohm2022attraction}. This "fragmentation" pattern will not change classification-based metrics as the small cluster will not affect the overall separation of classes. Thus, a new metric is proposed to evaluate this "fragmentation" pattern. A cluster ratio (Cluster-Ratio) metric is proposed as 
\begin{align}
Cluster-Ratio= \frac{1}{\exp^{\left\| Cluster_{hd} - Cluster_{ld} \right\|}}
\end{align} 
where $Cluster_{hd}$ and $Cluster_{ld}$ are the cluster numbers at the high dimension and low dimension, respectively. For low dimensional cluster numbers, OPTICS (Ordering Points To Identify the Clustering Structure) \cite{ankerst1999optics} is used to determine the number of clusters automatically. For high dimensional cluster numbers, it isn't easy to get accurate numbers from OPTICS due to the dimensionality. Thus, it is estimated from the labels. It should be noted that no detailed information is needed, and only the unique cluster numbers are extracted. The cluster number at the high dimension is considered unity (continuity assumption) for continuous numerical labels. If the cluster numbers at high dimension and low dimension are identified, a ratio of unity is obtained. The ratio decreases as the discrepancy of cluster numbers increases.

In summary, 14 metrics are proposed to evaluate different DR methods investigated in this study, covering local and local structure preservation. Table \ref{tab:metrics list} below lists all used metrics.

\begin{table}
    \centering
    \caption{List of used metrics for DR method evaluation}
    \begin{tabular}{|c|c|c|c|c|}
     \hline
        Name & Local or global & Required data & Ref & Other notes\\
        \hline
         Trust & both & $X$,$Y$ & \cite{lee2009quality} & co-ranking-based\\
         \hline
         Conti & both & $X$,$Y$ & \cite{lee2009quality} & co-ranking-based\\
         \hline
         LCMC & local & $X$,$Y$ & \cite{lee2009quality} & co-ranking-based\\
         \hline
         AUC & both & $X$,$Y$ & \cite{lee2015multi} & co-ranking-based\\
         \hline
         SVM-Classify & local & $X$,$Y$,$label$ & \cite{huang2022towards} & better for classification objective, need label\\
         \hline
         kNN-classify & local & $X$,$Y$,$label$ & \cite{huang2022towards} & better for classification objective, , need label\\
         \hline
         NPP & local & $X$,$Y$ & \cite{huang2022towards} & neighbor list intersection \\
         \hline
         Triplet & global & $X$,$Y$ & \cite{huang2022towards} & triple points evaluation \\
         \hline
         SpearCorr & global & $X$,$Y$ & \cite{huang2022towards} & rank correlation \\
         \hline
         Cen-kNN & global & $X$,$Y$,$label$ & \cite{huang2022towards} & centroid-based, need label \\
         \hline
         Cen-dist & global & $X$,$Y$,$label$ & \cite{huang2022towards} & centroid-based, need label \\
         \hline
         Curv-simi & local & $X$,$Y$ & proposed & evaluate curvature \\
         \hline
         NNWR & local & $X$,$Y$ & proposed & worst neighbor intersection\\
         \hline
         Cluster-ratio & global & $X$,$Y$, partial $label$ & proposed & only need number of label class\\
         \hline
    \end{tabular}

    \footnotesize{$^*$Name abbreviation can be found in the text; $X$ is the high dimension data, $Y$ is the low dimension embedding; $label$ is the label formation}
    \label{tab:metrics list}
\end{table}

\section{Experiments and results}
Experiments with eight benchmark datasets and five DR methods are performed compared with the 14 evaluation metrics mentioned above. Since most DR methods are proposed for unsupervised learning tasks. Only unsupervised learning comparison is shown in this section. Other learning settings and some parametric studies are performed and presented in the next section, "Discussions." The eight datasets used in \cite{wang2021understanding} include Swiss-Roll, Mammoth, coil-20, coil-100, MNIST, FMNIST, 20NG (20 newsgroups), and USPS. Details about these data are available from many available studies, including the one cited. The compared methods include tSNE, UMAP., TRIPMAP, PaCMAP, and CAMEL. All comparisons are made using Python 3.11.7 with the Anaconda (Mac Os 14.3.1, MacBook Pro with M1 Chip, 16 GB memory with 1TB hard drive). 

\subsection{Unsupervised Learning}
\begin{figure}
  \centering
  \includegraphics[scale=0.7]{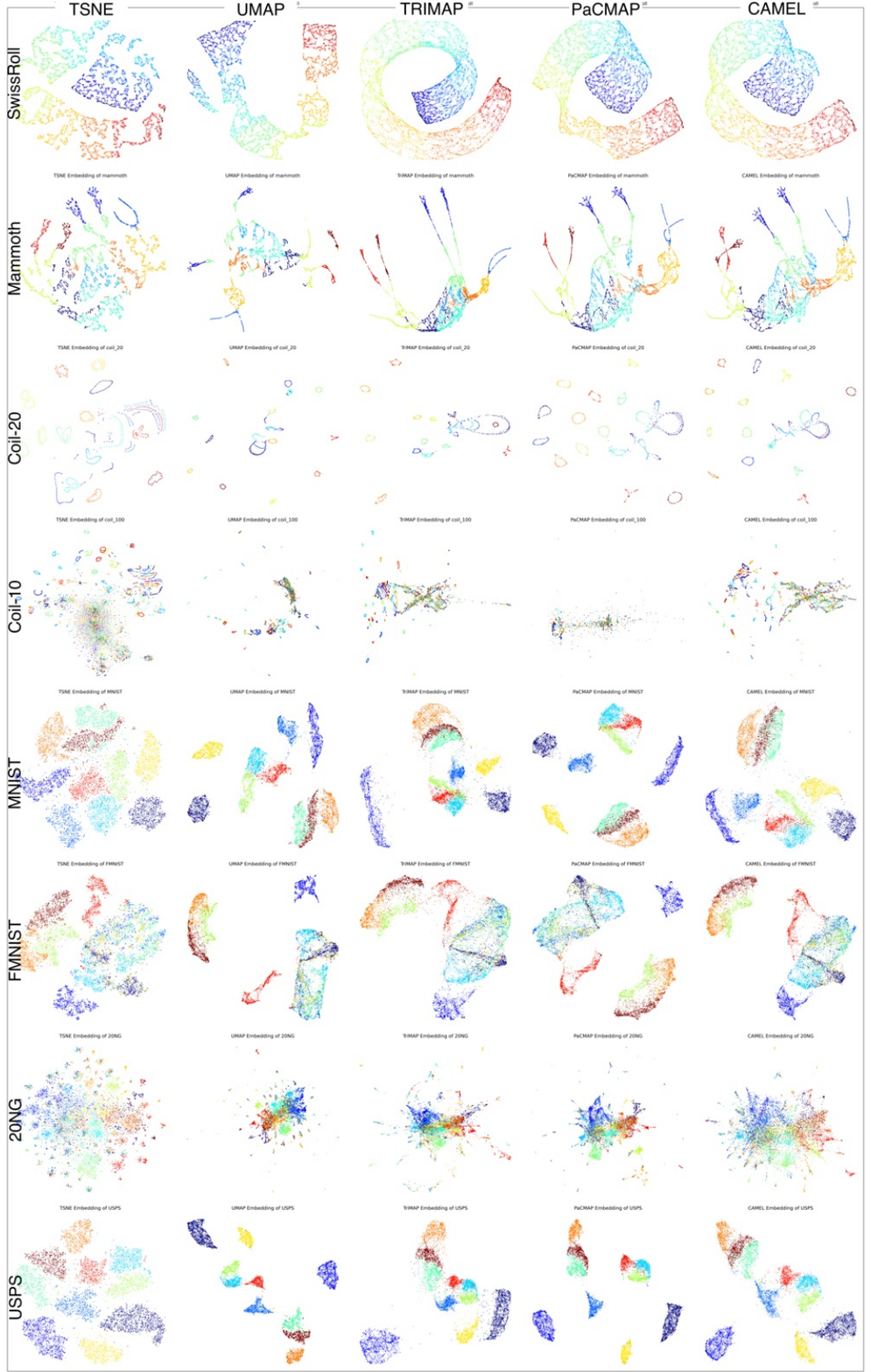}
  \caption{Visual comparison of 5 models with 8 datasets}
  \label{fig:model compare all}
\end{figure}

The first comparison is visual checking, as one of the main objectives of DR methods is to visualize the data in 2D or 3D.
The comparison of the five models with 8 datasets is shown in Fig. \ref{fig:model compare all}. All results are generated using the default setting in each model without further tuning. The data size is either the true data size or randomly sampled 10k points if the data size is beyond 10k. Swiss-Roll and Mammoth data are 3D toy data usually to see if the DR methods can preserve the 3D topology connectivity in 2D. It is observed that tSNE and UMAP seem to fragment the data, and TRIMAP, PaCMAP, and CAMEL can produce meaningful 2D projections of these 3D data. Coil-20, MNIST, FMNIST, and USPS data have similar trends for all 5 methods, as relatively structured clusters can be identified. tSNE appears to have more evenly distributed and not compact clusters, which has been explained in \cite{bohm2022attraction} due to the force spectrum of tSNE. UMAP, TRIMAP, PaCMAP, and CAMEL have similar clustering patterns, and the topological arrangement of clusters is also similar (ignoring the rotational variation). UMAP, TRIMAP, and PaCMAP have very compact smaller clusters for Coil-100 and 20NG datasets, and tSNE has well-spread clusters. CAMEL seems to be in between these patterns. Again, all these comparisons are purely visual, but all investigated methods appear to provide some reasonable embedding for the investigated datasets.  

\begin{figure}
  \centering
  \includegraphics[scale=0.6]{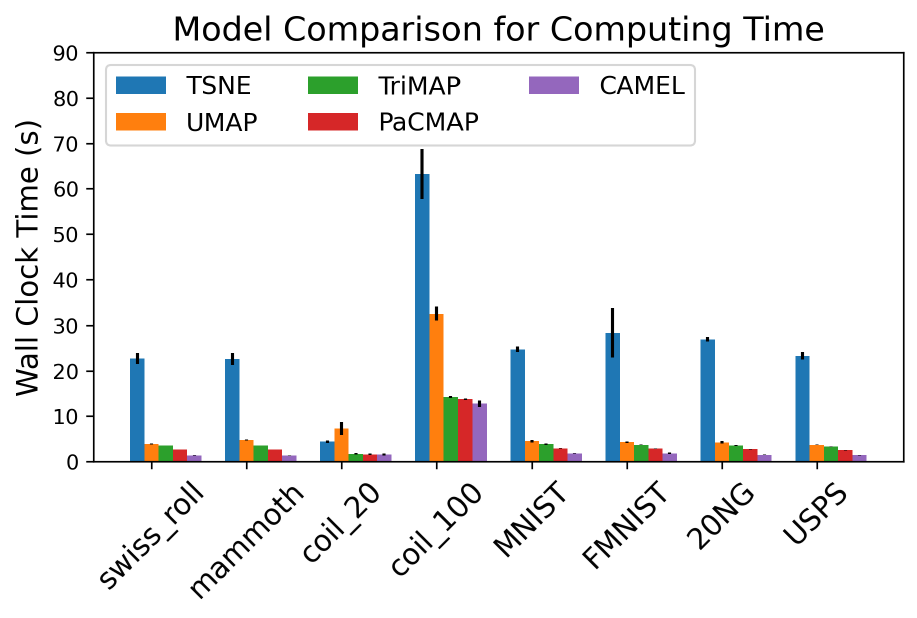}
  \caption{Computational time comparison for different DR models}
  \label{fig: computing time compare}
\end{figure}

A quick computational time comparison is performed before the detailed metrics are used for evaluation. The current computational is wall clock time comparison without detailed computational complexity analysis. Thus, this comparison also depends on the data pre-processing and code implementation. Nevertheless, we will use this comparison to show the difference from an end-user point of view. The difference with the error plot (one standard deviation from 5 runs) is shown in Fig. \ref{fig: computing time compare}. The proposed CAMEL is the most efficient model with a wall clock time speedup of 1.01X to 17.87X. Detailed values are presented in the appendix section as supplemental materials. PaCMAP is the second-fastest algorithm among the 5, and tSNE is generally the slowest algorithm for most datasets.

\begin{figure}
  \centering
  \includegraphics[scale=0.14]{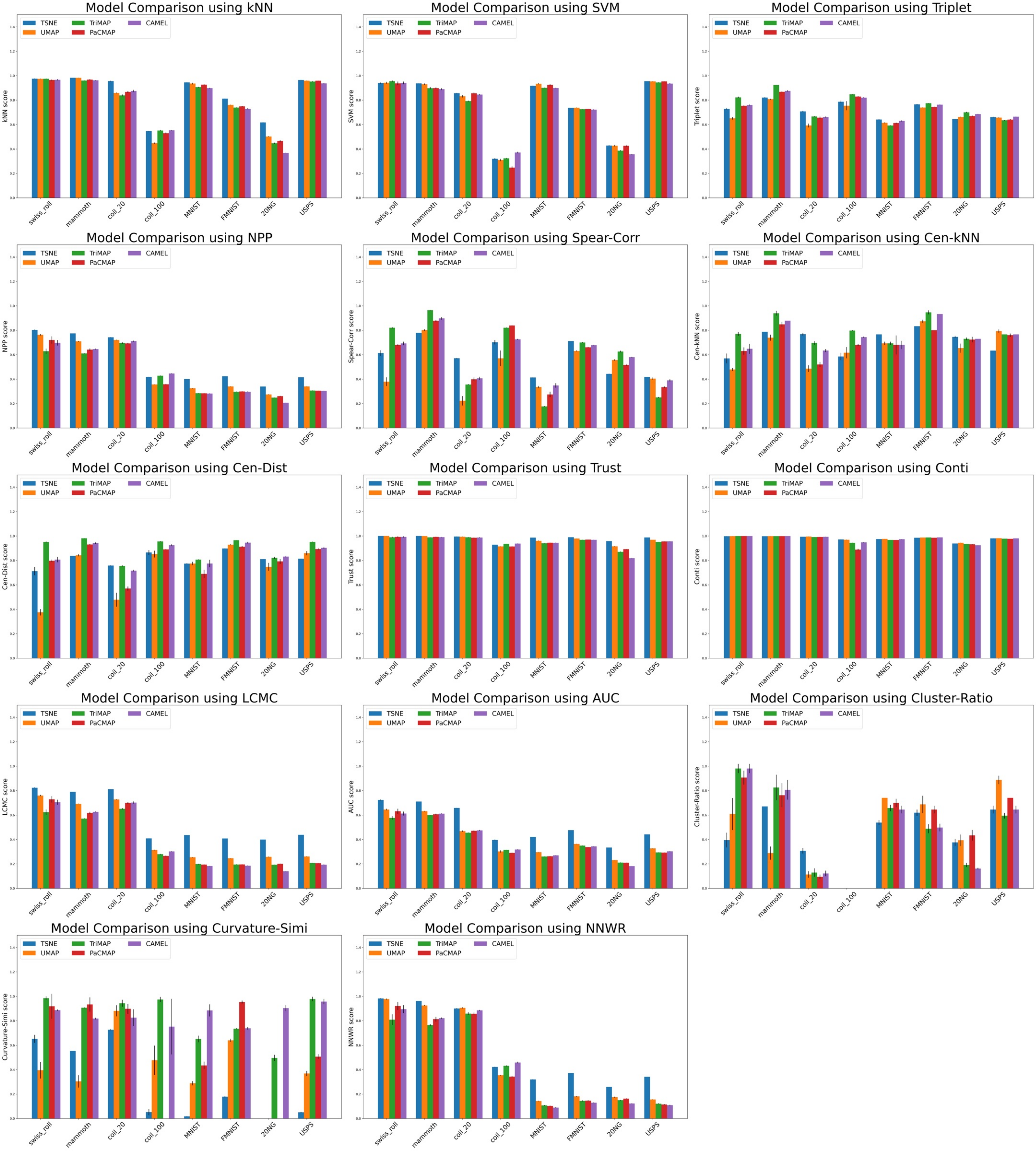}
  \caption{Metrics-based evaluation for different DR models}
  \label{fig: metrics evaluation of models}
\end{figure}

Finally, we will use the presented 14 metrics to evaluate all models for all used datasets. The results with error bars (one standard deviation from 5 runs) are shown in Fig. \ref{fig: metrics evaluation of models}. One can observe that Trust and Conti metrics show similar behavior for all models, and all scores are close to unity. This suggests that these two metrics do not have the power to separate different model performances (with the exception of 20NG using Trust). Very similar behavior can be found for the classification-based metrics (kNN and SVM). All models are almost equally good, except that 20NG datasets show better performance of tSNE as also shown in Trust and Conti. Similar behavior can be observed for other groups of local metrics, such as NPP, LCMC, AUC, and NNWR. tSNE has consistently the highest score in these metrics; UMAP has slightly higher scores than TriMAP, PaCMAP, and CAMEL, which are all very similar. It should be noted that NPP, LCMC, AUC, and NNWR are all derived using the neighbor point intersection between high and low dimensions. This suggests that the tSNE has the highest capability in terms of this metric. For global metrics, Triplet, Spear-Corr, Cen-kNN, and Cen-Dist show that all five models have some variation (small variation) for coil-20, coil-100, MNIST, FMNIST, 20NG, and USPS datasets, with TriMAP and CAMEL lead by a small margin. However, for Swiss-ROll and Mammoth datasets, these metrics consistently showed better performance for TriMAP, PaCMAP, and CAMEL compared to tSNE and UMAP. This is not very surprising as the metric Cluster-Ratio clearly shows that tSNE and UMAP misestimate the cluster numbers with a low score. This is because the fragmentation behavior of these models (see Fig. \ref{fig:model compare all}). UMAP leads in Cluster-Ratio metrics for MNIST, FMNIST, 20NG, and USPS datasets as it estimates the most accurate number of clusters. This is because the points tend to be more "concentrated" in UMAP, which helps the OPTICS algorithm to identify the correct number of clusters. Lastly, the Curvature-Simi metric shows that the proposed CAMEL is the most consistent and ranks higher. This is expected as CAMEL is the only model instrumented with explicit curvature computing. It is interesting that TriMAP also has a very good Curvature-Simi score. The author hypothesizes that TriMAP used triple points in addition to pairwise points as in other models. This approach includes some multi-body force interaction, as CAMEL has. In summary, it is very hard to select a single champion from all these comparisons of metrics. All investigated DR models excel in some metrics and lose in others. This is partially due to the different modeling mechanisms in different DR methods. If the objective is the classification with an unsupervised learning setting, all models should have very similar performance.    

\subsection{Supervised Learning}
\begin{figure}
  \centering
  \includegraphics[scale=0.6]{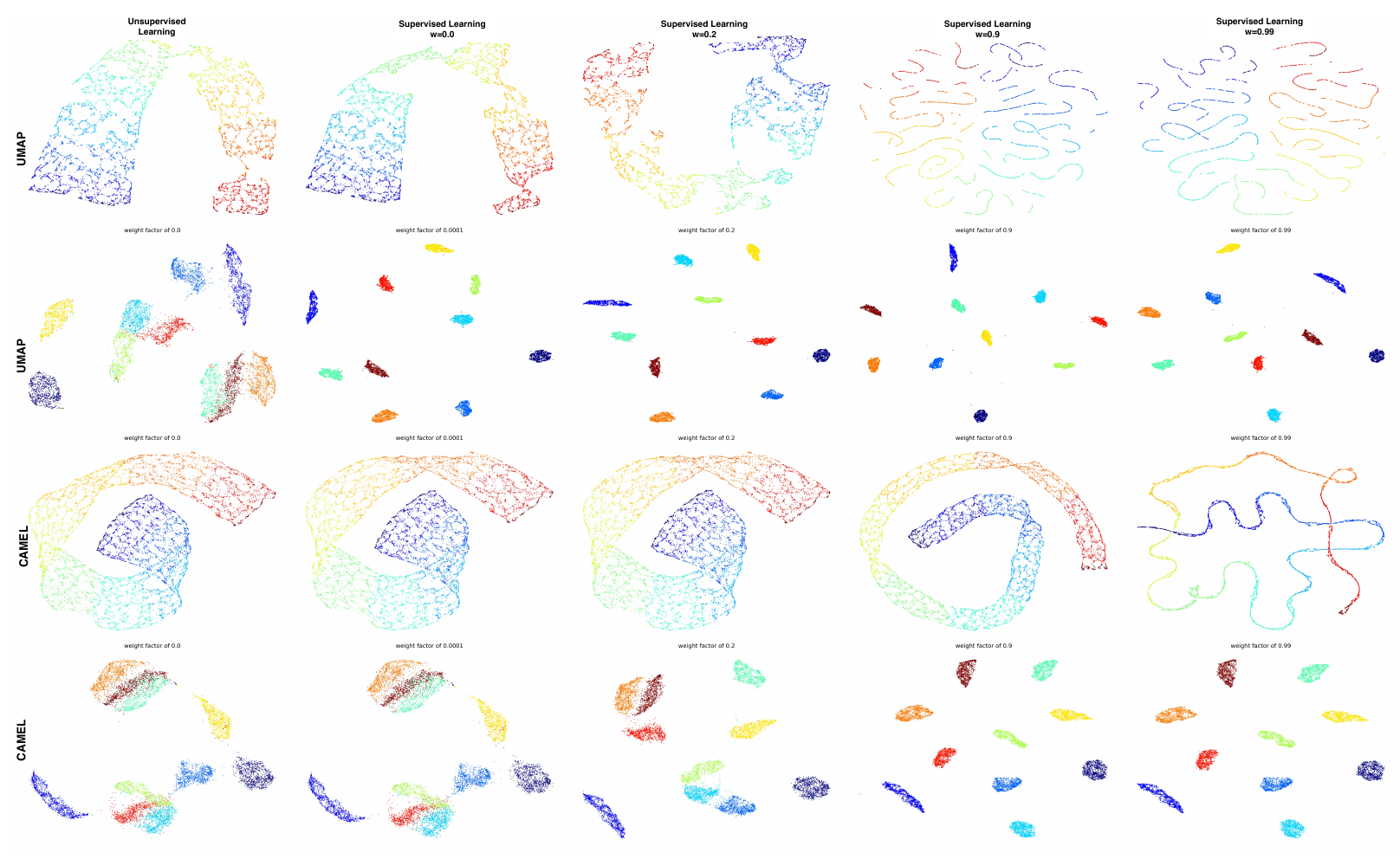}
  \caption{Comparison of UMAP and CAMEL for supervised learning}
  \label{fig: supervised learning compare}
\end{figure}

As discussed above, the proposed CAMEL is capable of supervised learning. Not many DR methods have developed supervised learning capability. UMAP has built-in supervised learning capabilities, although the way UMAP includes label information is very different than the proposed CAMEL. The two models are compared using Swiss-Roll and MNIST datasets, which cover both numerical and categorical label information. One hyperparameter in both models is the weight factor (from 0 to 1) of the label information. Thus, the comparison also checks the effect of different weight factors on the final embedding. The results are shown in Fig. \ref{fig: supervised learning compare}. The first two rows are UMAP results, and the last two rows are CAMEL results. The first column is the unsupervised learning results, and columns 2-5 have the weight factor of 0.0, 0.2, 0.8, and 0.99, respectively. Very different results are observed between the two models. First, UMAP did not reproduce the unsupervised learning when the weight factor is set to zero. The first and second column embedding are different. This is very apparent for the MNIST dataset. CAMEL reproduces the unsupervised learning when the weight factor is zero. Second, UMAP seems discontinuous when the weight factor changes from 0 to 1. In the Swiss-Roll data, UMAP produces fragmented line embedding when the weight factor is large. CAMEL has a smooth transition behavior. When the weight factor is 0.9, the narrowed roll embedding is observed, which makes sense as the label information starts to dominate and the original feature structure (3D coordinate in this case) becomes less important. When the extreme value of weight is 1 (0.99 in this example), the kNN only depends on the label information, and the problem reduces to 1D. Thus, a continuous rope is observed, which is the correct projection of the Swiss-Roll on one plane. Lastly, the current implementation of UMAP does not produce identical results even when the random state is set to a fixed value. In this experiment, the random state is set to 1 to remove the randomness effect on the comparison. CAMEL has consistent patterns for multiple runs. While UMAP did not have this behavior. This is likely not a theory issue, but rather the code implementation setting. In summary, the author believes CAMEL's supervised learning is beneficial due to the consistency, continuity, and repeatability of supervised embedding.    

\subsection{Metric Learning}
\begin{figure}
  \centering
  \includegraphics[scale=0.7]{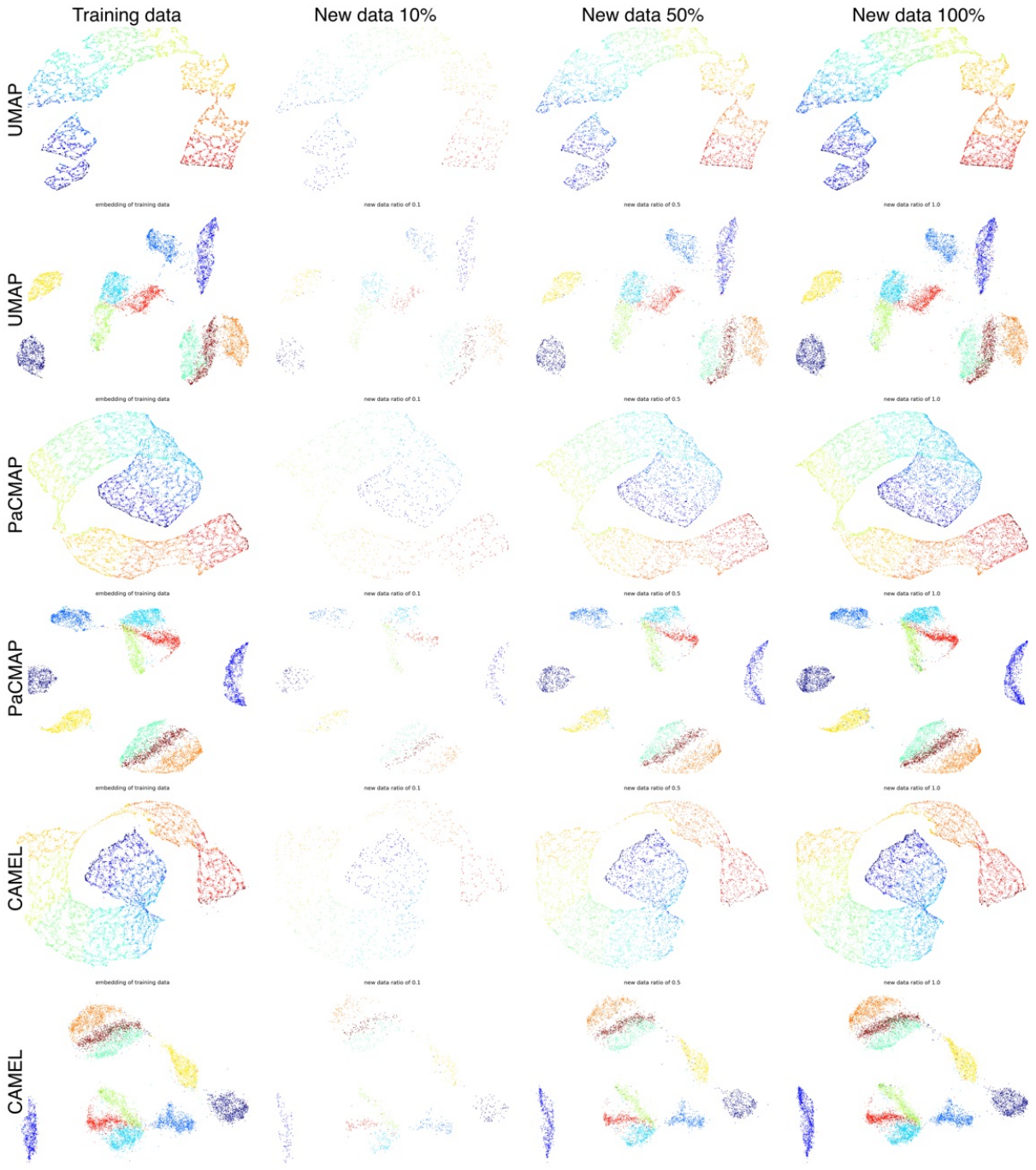}
  \caption{Comparison of UMAP, PaCMAP, and CAMEL for unsupervised metric learning}
  \label{fig: unsupervised metric learning compare}
\end{figure}

As discussed in the algorithm section. The "Metric Learning" in the current paper is defined as projecting new testing data to the embedding obtained from the training data. This projection can be in unsupervised mode or supervised mode. The key idea is to learn a metric mapping for any future data points. UMAP, PacMAP, and CAMEL are all instrumented with this metric learning in unsupervised learning mode. The results for the three models are shown in Fig. \ref{fig: unsupervised metric learning compare}. The first column is the embedding plot from the training data. The second to the fourth columns are the embedding plots for new data with $10\%$, $50\%$, and $100\%$ of the training data size. As can be observed, all three models can successfully project new points to a stable embedding space.

\begin{figure}
  \centering
  \includegraphics[scale=0.55]{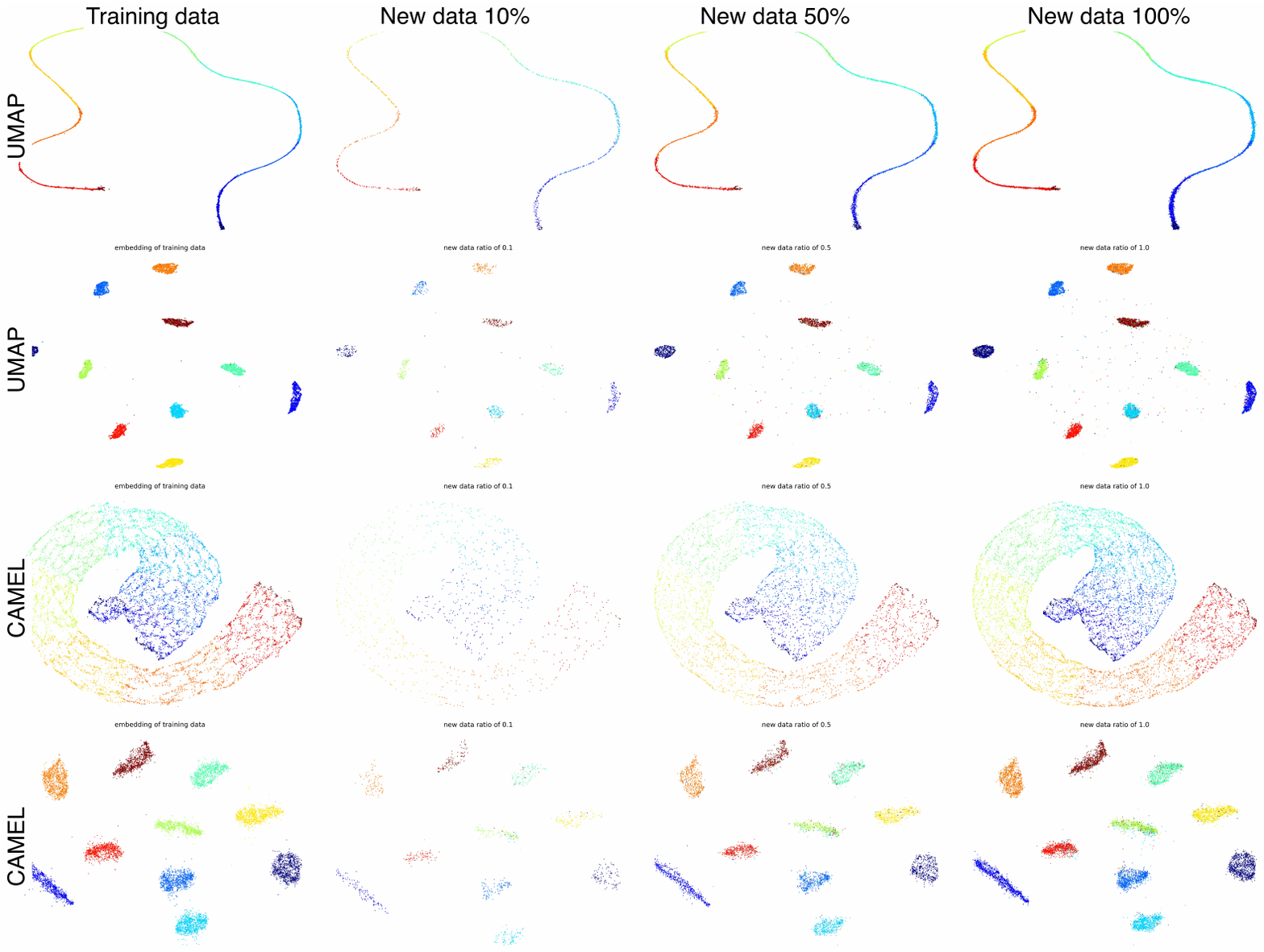}
  \caption{Comparison of UMAP and CAMEL for supervised metric learning}
  \label{fig: supervised metric learning compare}
\end{figure}

Only UMAP and CAMEL can perform supervised metric learning and results are shown in Fig. \ref{fig: supervised metric learning compare}. All parameters are using the default setting. Again, both models can successfully project new data points to the same embedding space in the supervised learning mode.

\subsection{Semi-supervised learning}
\begin{figure}
  \centering
  \includegraphics[scale=0.12]{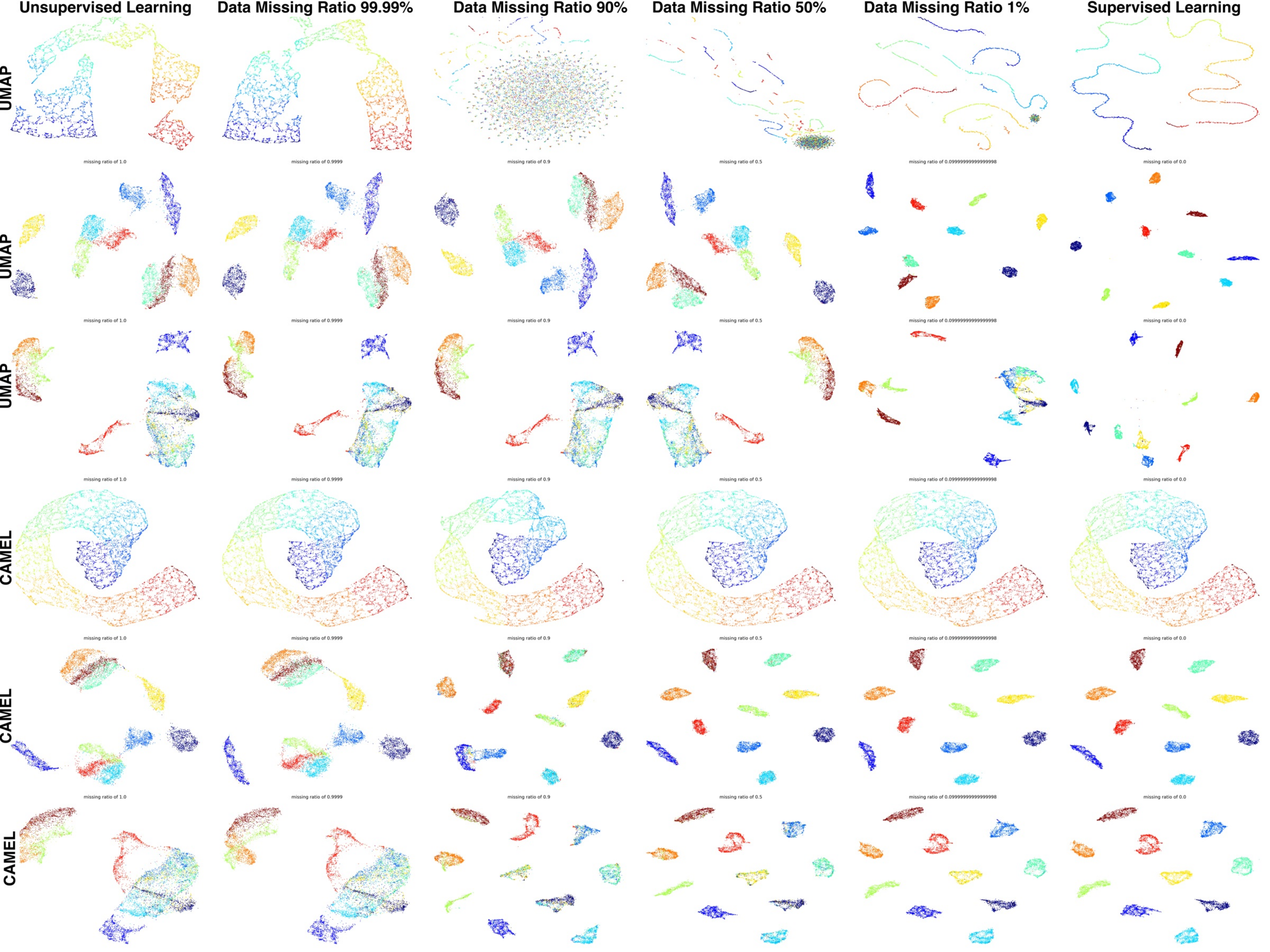}
  \caption{Comparison of UMAP and CAMEL for semi-supervised learning}
  \label{fig: semi-supervised learning compare}
\end{figure}

Semi-supervised learning is another widely used ML method in which the data missing ratio is critical. In general, if the method can achieve very good performance with a small percentage of labeled data, that is very beneficial from a practical point of view. UMAP and CAMEL have the capabilities of semi-supervised learning and are compared. Swiss-Roll, MNIST, and FMINST are used for comparison. The results are shown in Fig. \ref{fig: semi-supervised learning compare}. The first three rows are UMAP results and the last three rows are CAMEL results. The first column is the unsupervised results and the last column is the supervised learning results. Columns 2-5 have the data missing ratio of $99.99\%$, $90\%$, $50\%$, and $1\%$, respectively. The missing ratio of $99.99\%$ should approach the unsupervised learning and the missing ratio of $1\%$ should approach the supervised learning.  It is observed that UMAP and CAMEL have very different behaviors in terms of semi-supervised learning. First, CAMEL archives perform very well with minimal labeled data. For the investigated datasets,  $10\%$ additional labeled data is already able to construct a stable well-separated cluster. This can be seen for the FMNIST datasets, as it is seen in the previous section that unsupervised learning cannot achieve well-separation for classification jobs. It a little surprising to see that UMAP did not achieve the well-separation even with the $99\%$ of the labeled data. Second, CAMEL archives a smooth and stable transition from unsupervised learning to supervised learning as the available label information increases. This is clearly seen in all three datasets. UMAP, on the other hand, does not have this smooth and stable behavior. MNIST and FMINIST embedding seems to vary a lot even the random state is fixed. Finally, UMAP appears to have difficulties with continuous numerical labels, as the Swiss-Roll data has strange embedding behavior with partial labels. For continuous Swiss-Roll data, unsupervised and supervised learning have similar embedding using CAMEL. Thus, semi-supervised learning has almost identical stable results irrespective of the missing data ratio. UMAP seems to diverge for semi-supervised learning when the missing data ratio is in the middle range of 0-1. It is shown that the proposed CAMEL appears to have a better semi-supervised learning performance.

\subsection{Inverse Learning}
\begin{figure}
  \centering
  \includegraphics[scale=0.25]{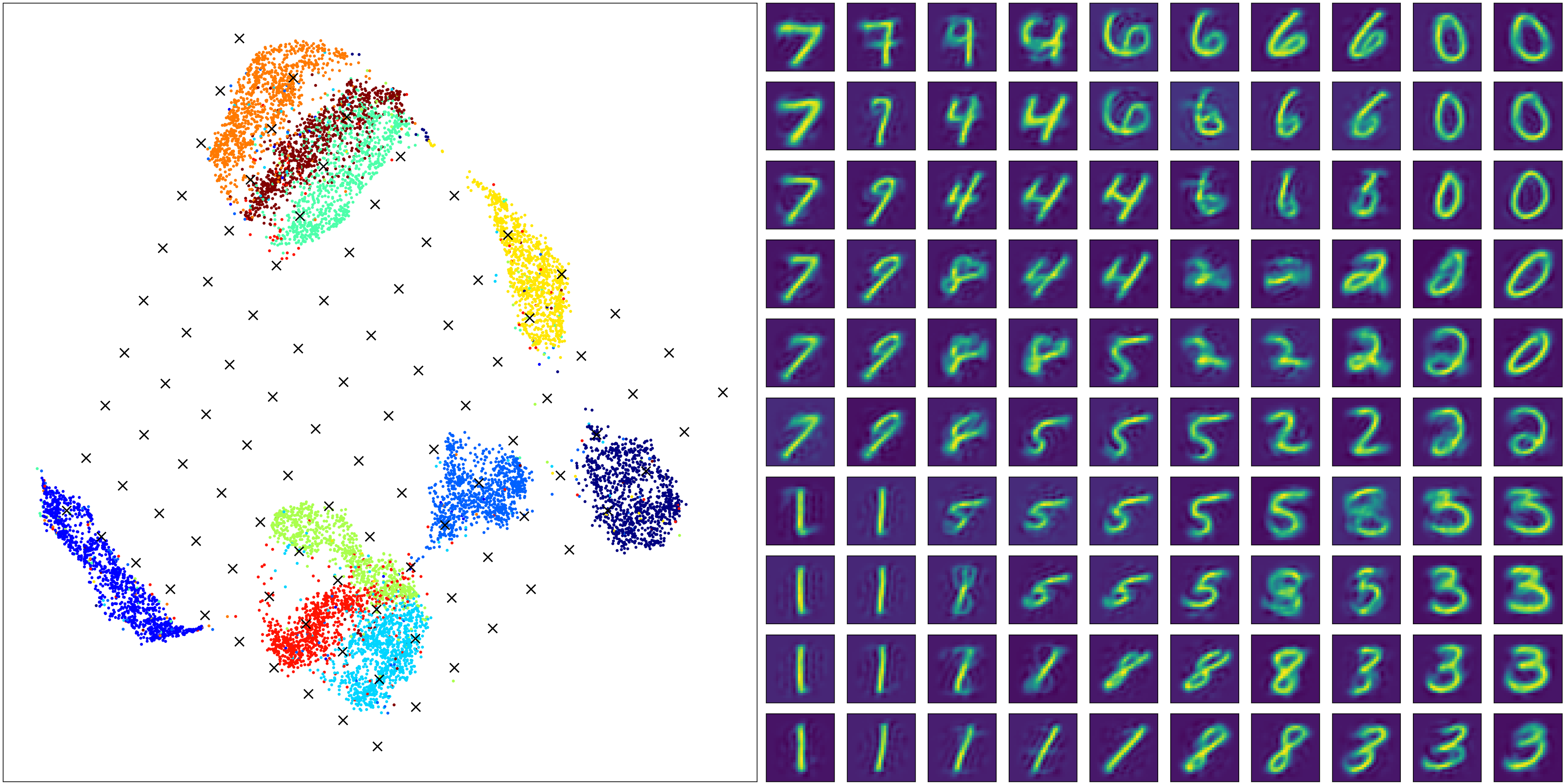}
  \caption{Inverse learning for MNIST data using CAMEL - Interpolation Initialization}
  \label{fig: inverse learning camel interpolate}
\end{figure}

One of the most interesting applications of DR is the generative modeling from low dimension to high dimension. This can be considered as the inverse of dimension reduction (i.e., dimension augmentation), which is in analogy to the well-known decoder structure in neural networks. The proposed CAMEL used the same method without introducing new components as the proposed force field is applied to both high-to-low and low-to-high projection. Inverse learning is first applied to MNIST datasets. 100 grid points (10x10) are selected in the low-dimension embedding, and corresponding projections to the high-dimension space are visualized as pictures on the right. The grid points and images on the right follow the same ordering, so the correspondence of high and low dimensions can be investigated. The results are shown in Fig. \ref{fig: inverse learning camel interpolate}. It is observed that the inverse correctly reproduces the original category characteristics and also shows the topological transition between categories. For example, rows 1-3 and columns 2-3 (top left corner) show the topological similarities of digits 7, 9, and 4. They form nearly parallel boundaries in low dimensions, and high-dimensional images also show strong correspondence along this boundary. Another example is on rows 7-8 and columns 7-8 (the triple junction in the bottom right corner. This triple junction is not well separated in the low dimensional space and the reason is that the digits 5, 8, and 3 have high similarity in the high dimensional image space. Thus, the proposed CAMEL inverse learning appears to capture the main topological features in the image. 

\begin{figure}
  \centering
  \includegraphics[scale=0.25]{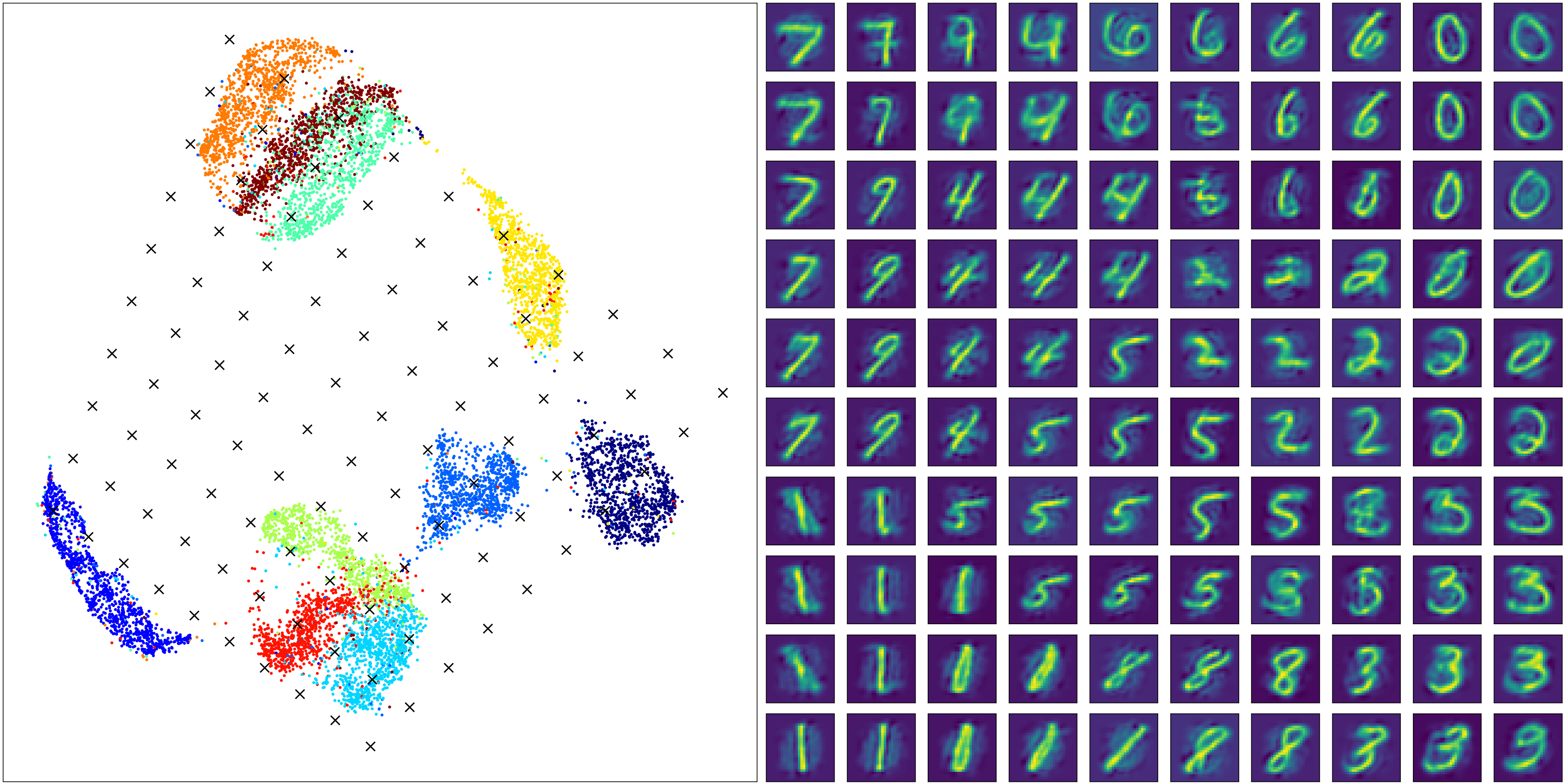}
  \caption{Inverse learning for MNIST data using CAMEL - Random Initialization}
  \label{fig: inverse learning camel random}
\end{figure}

In order to illustrate that this inverse learning is not determined by the initialization, a "random" initialization is used to produce the same images using CAMEL, and the results are shown in Fig. \ref{fig: inverse learning camel random}. It is shown that the random initialization can also produce similar inverse learning images, with more blurriness and noise in the images. Thus, interpolation initialization is suggested for inverse learning if the constructed low-dimensional space is stable and of high quality.

\begin{figure}
  \centering
  \includegraphics[scale=0.25]{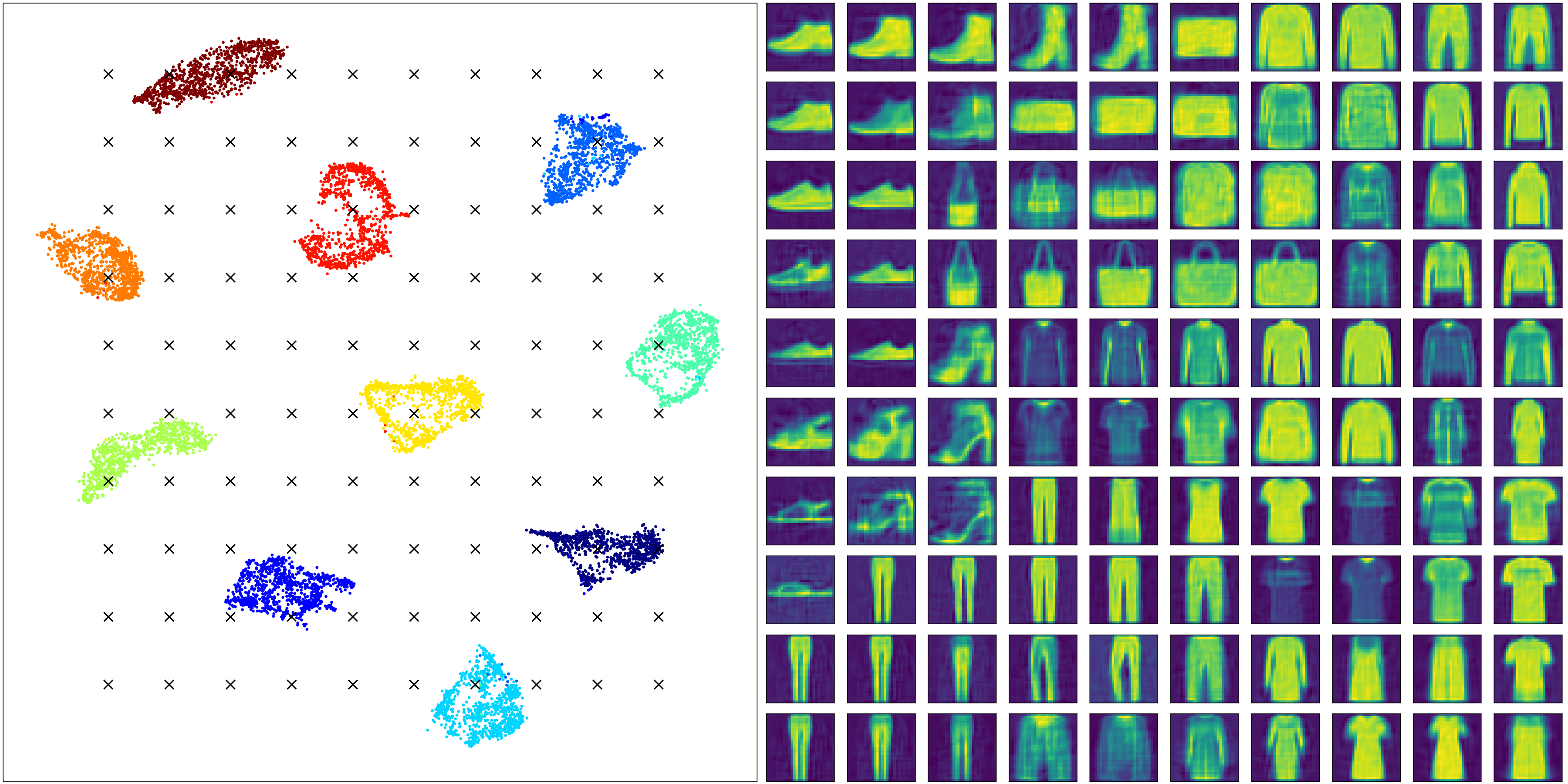}\\
    \vspace{0.5cm}
  \includegraphics[scale=0.25]{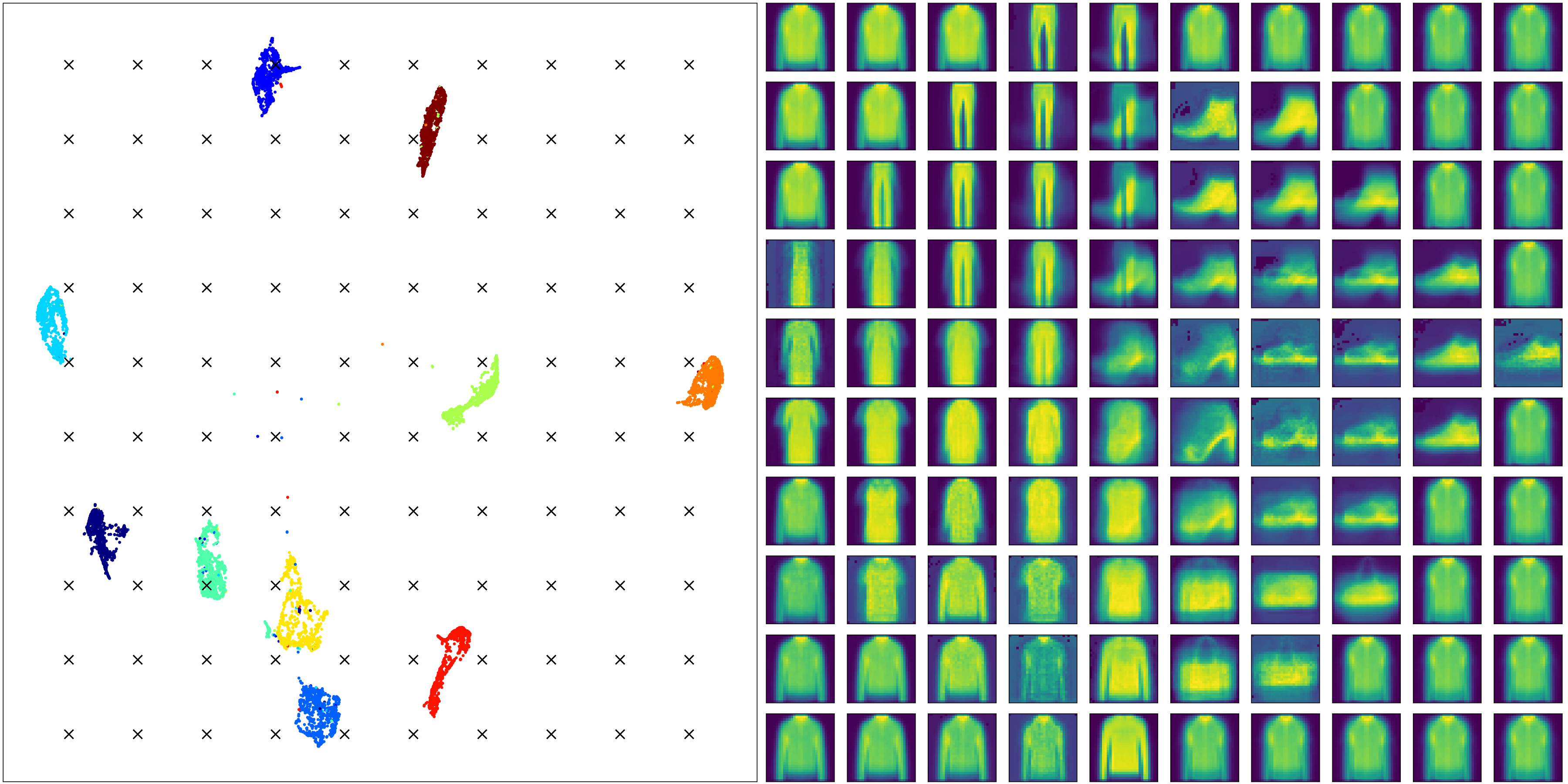}  
  \caption{Inverse learning for FMNIST data using CAMEL (top) and UMAP (bottom)}
  \label{fig: inverse learning camel umap compare}
\end{figure}

Not many DR methods are instrumented for inverse learning. UMAP has this capability, and the author thanks its template for its image-based evaluation of inverse learning capability. A comparative study is performed using the FMNIST dataset. The low-dimensional embedding is constructed using the supervised learning setting, as shown above. The comparison results are shown in Fig. \ref{fig: inverse learning camel umap compare}. It is seen that UMAP (bottom figure) has a very compact localized cluster but has less local structure information within a cluster. The inversely generated images by UMAP have less variety and cleanness, both of which are critical for generative models as the details are very critical. The proposed CAMEL also has well-separated clusters (i.e., good for classification), but has more local structure within the clusters. The inversely generated images have much more details and cleanness (see the laces in shoes and handles in bags). It also shows some true generative power as the top right corner image and bottom row fourth column image never appeared in the training datasets, which are "novel" categories to the FMNIST datasets. Thus, CAMEL appears to have a better generative power and inverse construction capability than its peers.

\section{Discussions}
This section focuses on investigating the parameter effects of CAMEL on the final performance of embedding, which are presented using parametric studies of several parameters. 
\subsection{Effect of curvature force weight coefficient}
\begin{figure}
  \centering
  \includegraphics[scale=0.18]{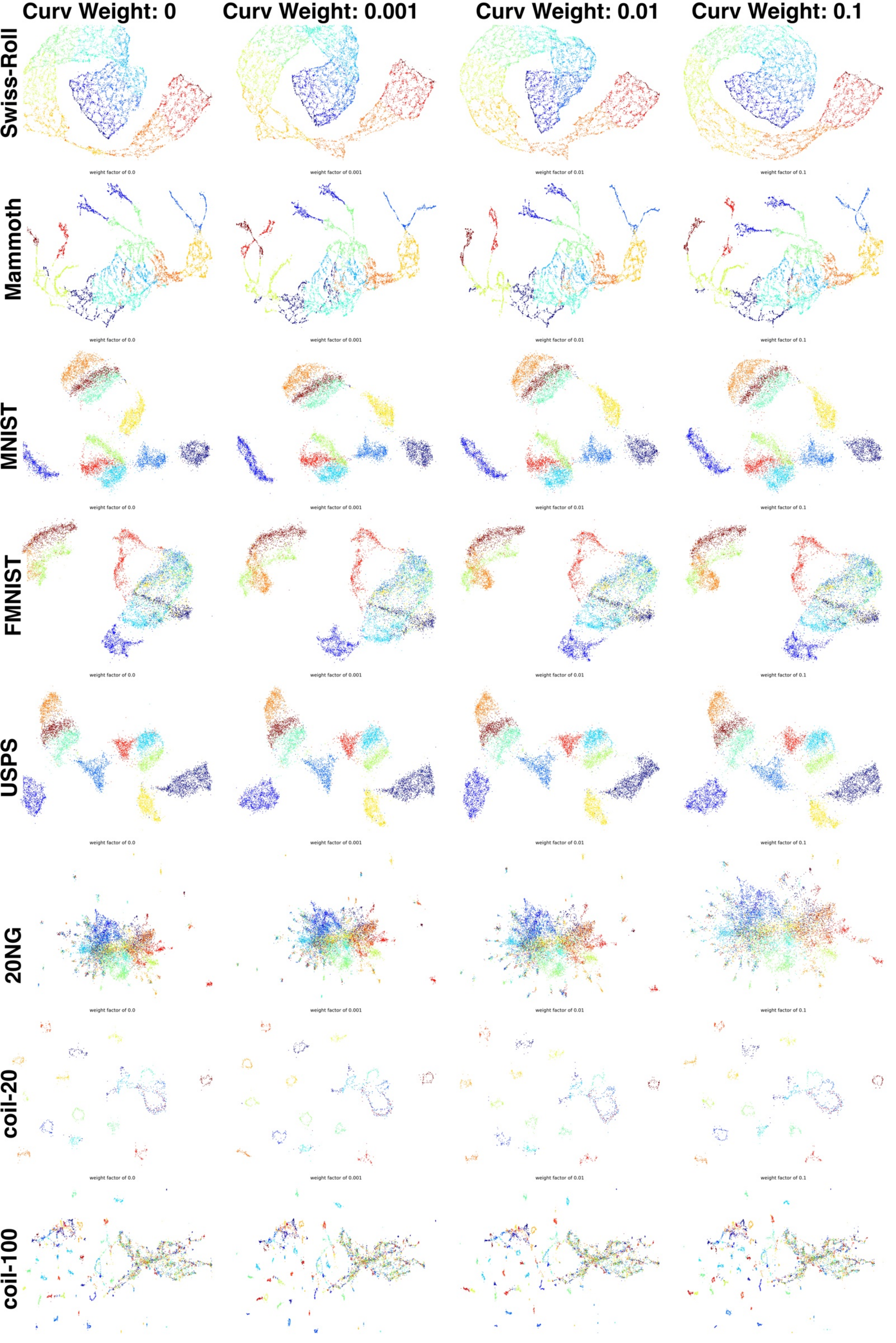}
  \caption{Embedding of CAMEL using different curvature weight coefficients}
  \label{fig: camel different curvature weights}
\end{figure}

\begin{figure}
  \centering
  \includegraphics[scale=0.8]{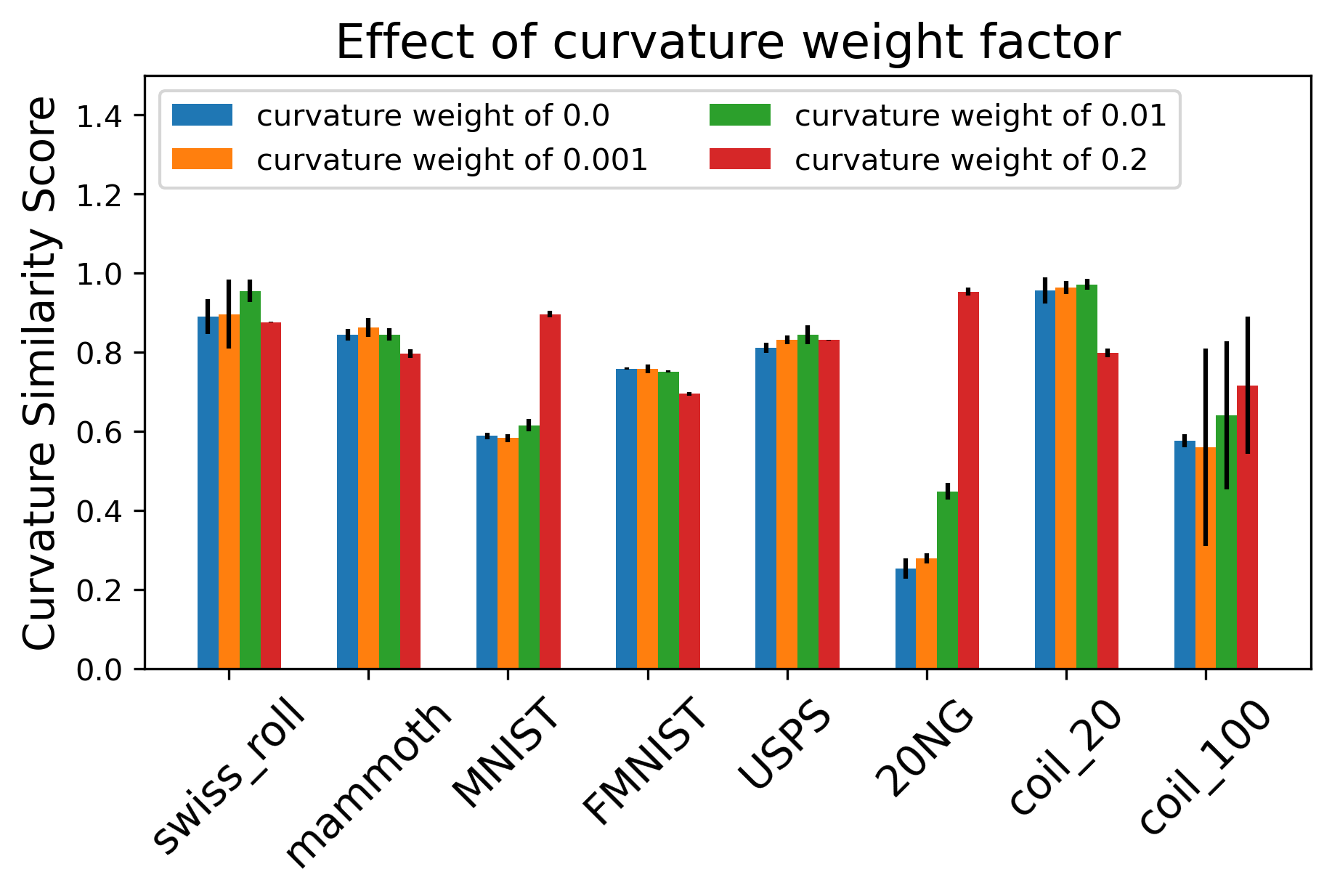}
  \caption{Curvature Similarity Score of CAMEL using different curvature weight coefficients}
  \label{fig: curvature metrics with curvature weights}
\end{figure}

One of the most unique contributions of the proposed CAMEL is to include curvature information in the force field computation. Since the curvature only adds additional force to the neighbor's attractive force and distant points' repulsive force, it is expected that the major global structure will not be affected by the curvature-induced force. Instead, it will only augment the neighbor point attractive force and affect the local structure. Thus, we named it Curvature-Augmented Manifold Embedding and Learning (CAMEL). To verify this hypothesis, we change the curvature weight coefficient in the range $0.0, 0.001, 0.01, 0.2$. The curvature similarity score is computed together with the embedding visualization. The embedding results are shown in Fig. \ref{fig: camel different curvature weights}. It is observed that the global structure does not change much for different weight factors. Small local structure change for 20NG and MNIST is observed. The curvature similarity score is shown in Fig. \ref{fig: curvature metrics with curvature weights}. It is seen that the curvature similarity score can increase with weight coefficient increase (20NG, coil-100, and MNIST) or increase/remain stable until 0.01 and drop slightly at 0.2(other datasets). This suggests that the effect of curvature weight coefficients seems to depend on specific datasets. 0.01-0.1 seems to be the good estimated values based on the current investigation.  

\subsection{Effect of number of neighbors}
\begin{figure}
  \centering
  \includegraphics[scale=0.3]{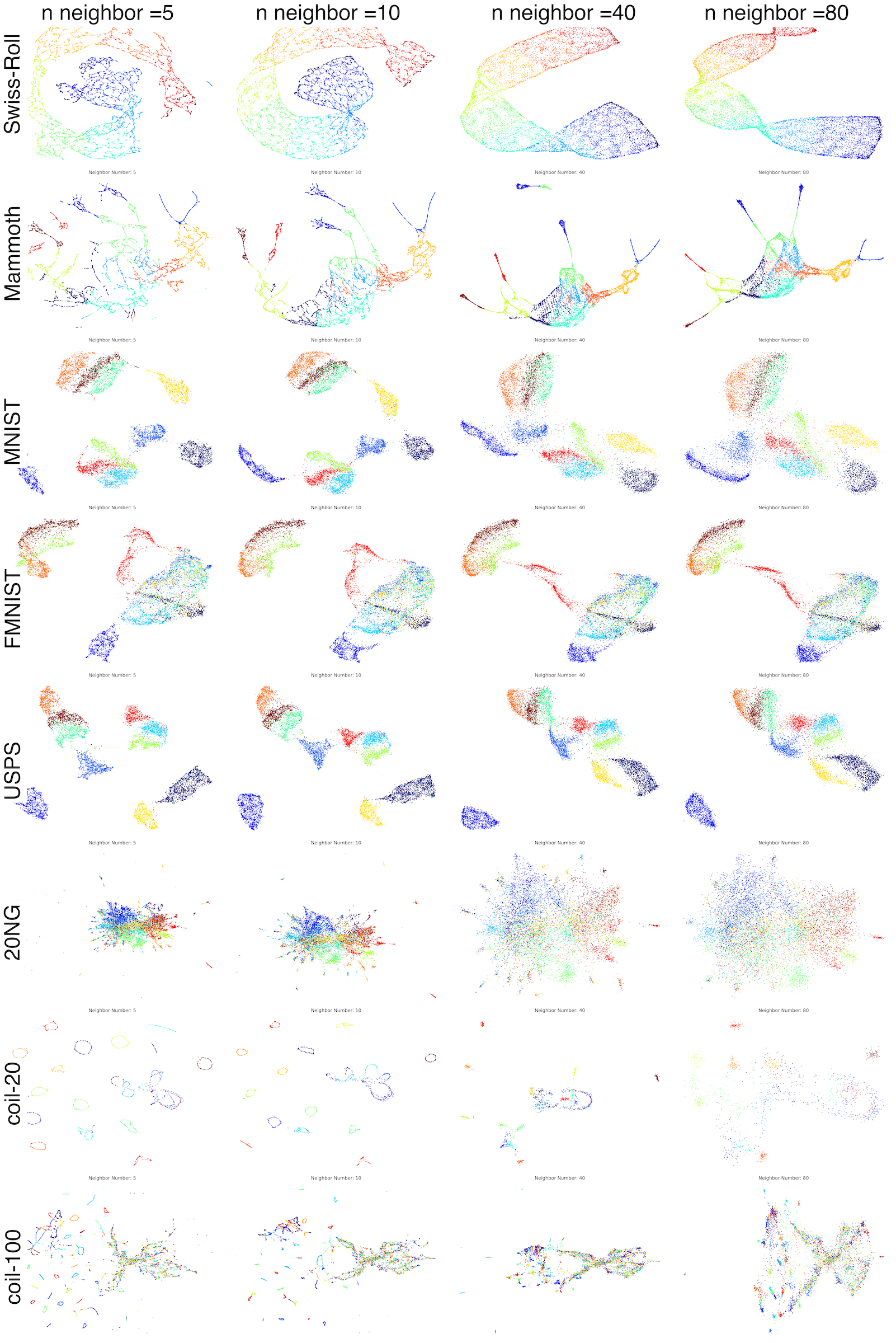}
  \caption{Embedding of CAMEL using different neighbor numbers with proposed modification}
  \label{fig: neighbor effects with modification}
\end{figure}

\begin{figure}
  \centering
  \includegraphics[scale=0.3]{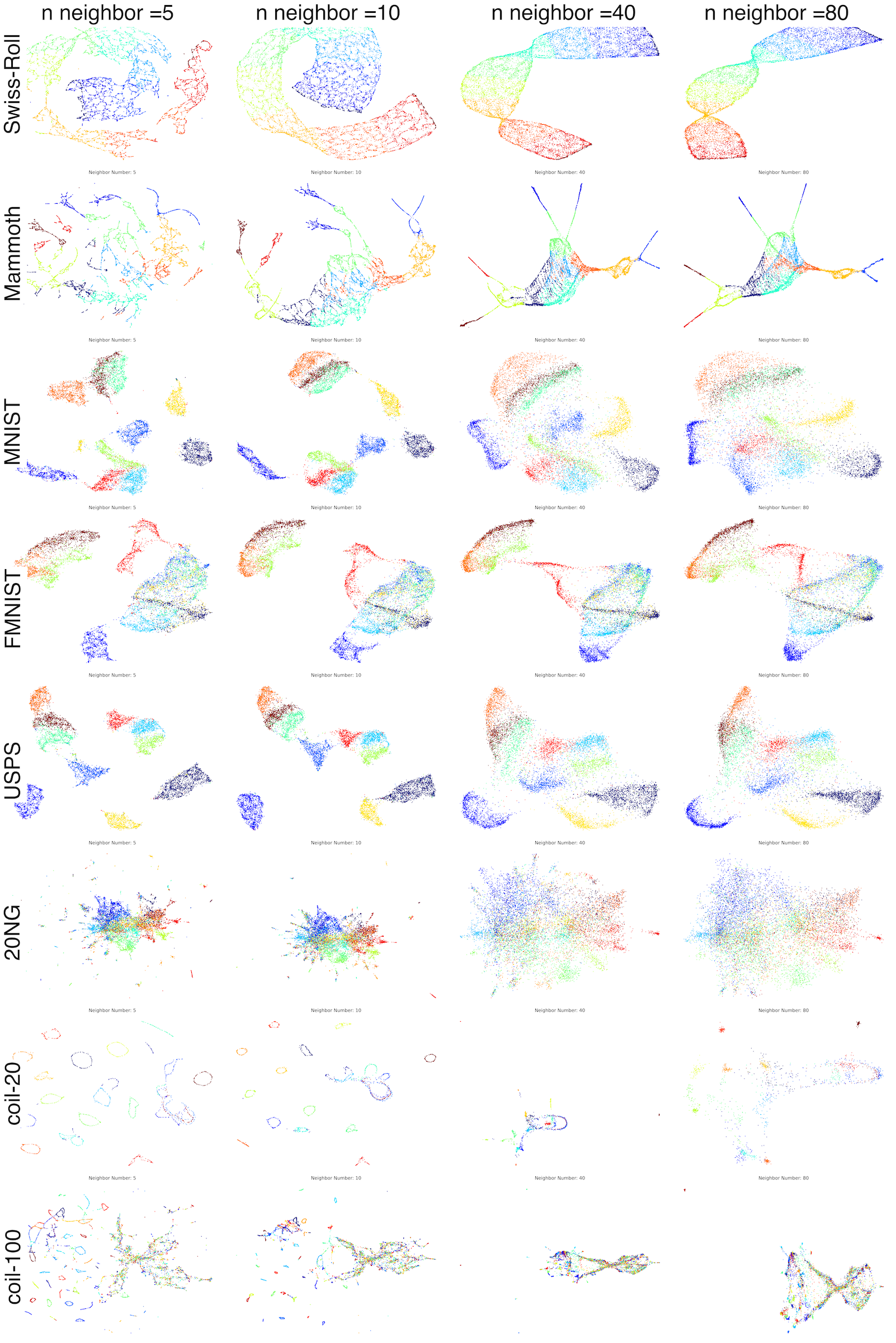}
  \caption{Embedding of CAMEL using different neighbor numbers without proposed modification}
  \label{fig: neighbor effects without modification}
\end{figure}

As discussed in the modeling section, the proposed CAMEL incorporates the number of neighbors and negative sampling ratio (distant points) in the force field weight coefficient calculation. Thus, the number of neighbors and distant points in ideal conditions should not affect the final results. Results for embedding with the proposed neighbor number modification are shown in Fig. \ref{fig: neighbor effects with modification}. It is seen that the main structure remains stable for different numbers of neighbors, with the exception of a few, such as 20NG. This is more apparent when compared with the embedding without the proposed modification, as shown in Fig. \ref{fig: neighbor effects without modification}. It is seen that the embedding has a relatively large variation with a different number of neighbors. In conclusion, the proposed modification of weight coefficients with the number of neighbors is effective but does not fully eliminate the results' dependency on the number of neighbors. Further study is required to explain the number of neighbor point effects.

\subsection{Effect of Negative Sampling}
\begin{figure}
  \centering
  \includegraphics[scale=0.3]{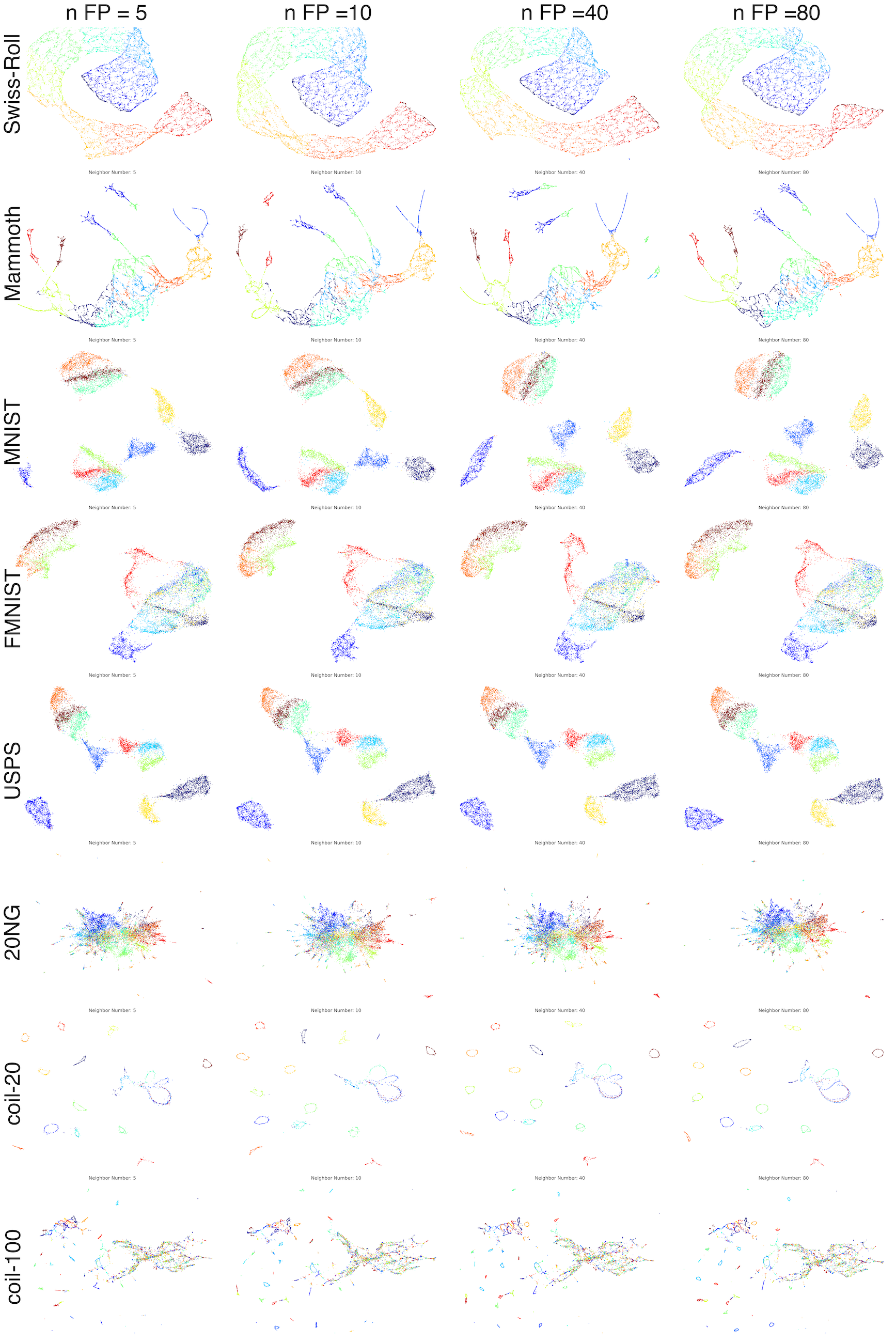}
  \caption{Embedding of CAMEL using different negative samples with proposed modification}
  \label{fig: FP effects with modification}
\end{figure}

\begin{figure}
  \centering
  \includegraphics[scale=0.7]{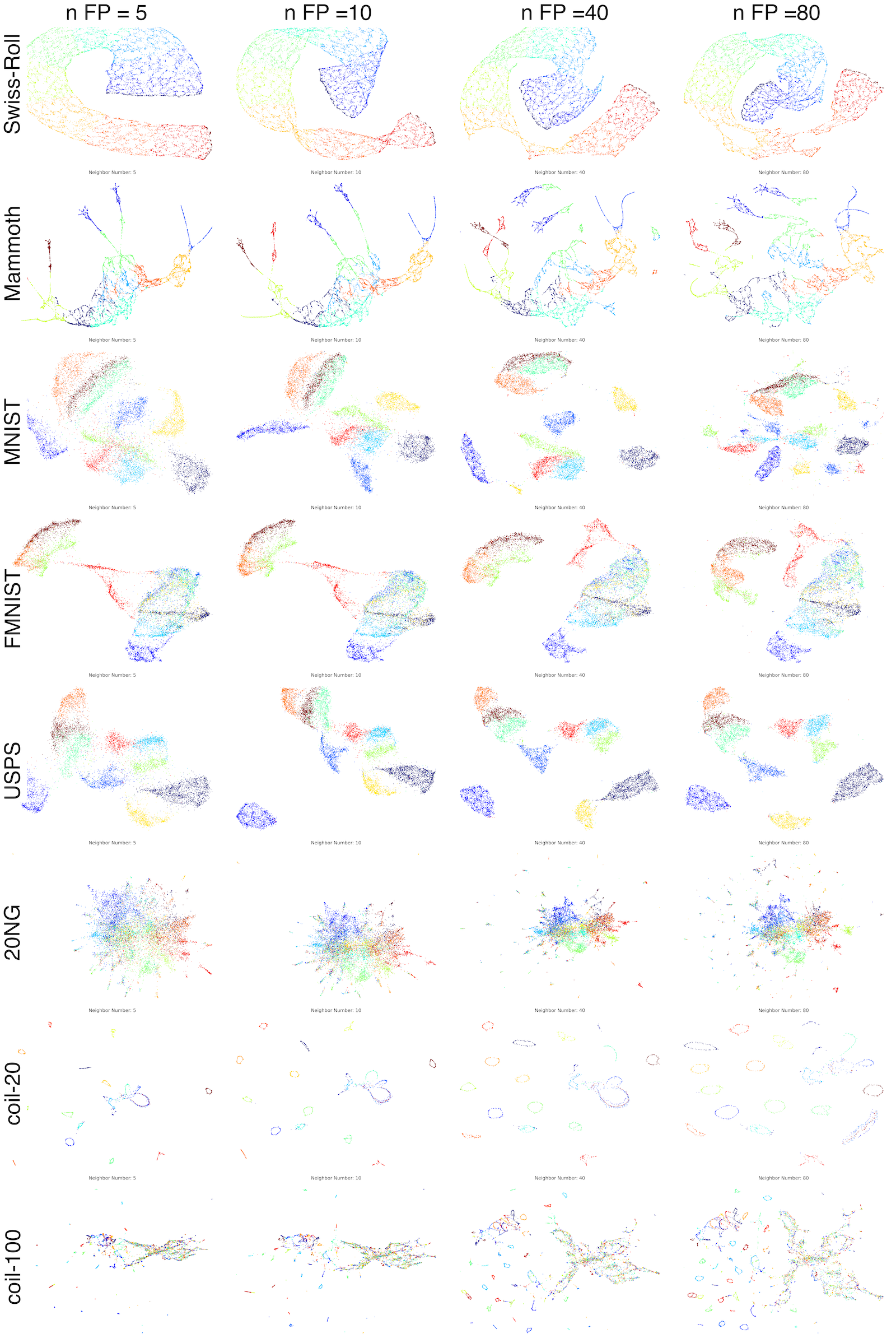}
  \caption{Embedding of CAMEL using different negative samples without proposed modification}
  \label{fig: FP effects without modification}
\end{figure}

Negative samples are used for the repulsive force, and different samples of negative samples can be used. The proposed CAMEL modifies the repulsive forces coefficient according to the number of negative samples. The results of embedding with the proposed modification are shown in Fig. \ref{fig: FP effects with modification}. It is clearly seen that the embedding is very stable, irrespective of the negative sample numbers. This is very apparent when compared with the embedding without the proposed modification, as shown in Fig. \ref{fig: FP effects without modification}. The embedding becomes very different when many negative samples are used. This clearly demonstrates that the negative sample should be considered as a model setting rather than a parameter to tune. 

\section{Conclusion}
A novel dimension reduction (DR) and data visualization model, CAMEL, is proposed. CAMEL stands for Curvature-Augmented Manifold Embedding and Learning, which belongs to the general DR methods using the attractive and repulsive force fields to find the embedding. The unique contributions of the proposed CAMEL are:
- Include the force caused by the curvature using a multibody potential, inspired by the particle dynamics in physics. The curvature force augments the neighbor's attractive force and far points' repulsive force, enabling local curvature mapping between high and low dimensions.
- Include modification of force weights to explicitly include the effect of neighbor numbers and negative sample numbers, which reduces the trial-and-error calibration of model parameters. This is achieved by the force equilibrium conditions in the physics/mechanics modeling.
- Propose new metrics for DR method evaluation, particularly on the curvature metrics, which match the curvature similarity between high and low dimensional spaces. Another new metric is the automatic cluster retaining ratio, which focuses on the fragmentation behavior in dimension reduction operations.
- Develop new algorithm implementation for supervised learning by including the modification of weighted kNN graph using label information. 
- Develop a new algorithm implementation for metric learning using partial freezing force field computing to map new testing points in the embedding space.
- Develop a new algorithm implementation for semi-supervised learning by performing kNN imputation of unseen/missing labels within CAMEL. The results show a consistent and smooth transition between unsupervised and supervised learning. It has also been demonstrated that a small percentage of labels can stabilize the final embedding compared with the full label cases.
- Develop a new algorithm for inverse embedding (dimension augmentation and generative learning) by inversely applying the force field in high dimensional space. The results show good variety and clear details for the inverse embedding, compared with the state-of-the-art DR methods.
- The computational efficiency of the proposed CAMEL is superior to its peers with a typical speedup factor of 1x-20x, depending on the datasets

The author believes that the proposed CAMEL offers many benefits for DR and data visualization. However, the author also realizes a few major research directions for the proposed CAMEL.

- The current CAMEL is a pure data-driven non-parametric model. Integration with other parametric and non-parametric models, such as Neural Networks and Gaussian Processes, can be very useful.
- One of the powerful uses of CAMEL is the automatic and robust feature extraction, which can be used as an upper-stream processing tool for many models requiring dimension reduction, such as Reinforcement Learning. Concurrent implementation in this regard needs further study.
- One critical element missing in the current study is self-supervised learning. The developed algorithms can help to realize a set of self-supervised learning task designs. This is ongoing work in the continued CAMEL development.

\section*{Acknowledgments}
 The research is partially supported by funds from the National Science Foundation (Award No. 2331781). The support is greatly acknowledged. 

\section*{Code Availability}
All codes and reproducible data can be accessed at https://github.com/ymlasu/CAMEL, and related documentation can be found at https://camel-learn.readthedocs.io/en/latest/

%Bibliography
\bibliographystyle{unsrt}  
\bibliography{references}

\end{document}